%% file: main.tex
\PassOptionsToPackage{no-math}{fontspec}
\documentclass[11pt]{article}
\usepackage{fontspec}
\usepackage{authblk}
\usepackage[a4paper,margin=1in]{geometry}
\usepackage{graphicx} 
\usepackage{amsmath}
\usepackage{booktabs}
\usepackage{placeins}
\usepackage{amssymb,mathrsfs}
\usepackage{array,longtable,tabularx,adjustbox,fvextra,xurl,newunicodechar}
\usepackage[authoryear,round]{natbib}
\makeatletter
\renewcommand{\bibpreamble}{%
  \let\@biblabel\NAT@biblabelnum
  \let\@bibsetup\NAT@bibsetnum
}
\makeatother
\usepackage[unicode,hidelinks]{hyperref}
\allowdisplaybreaks
\newcommand{\tightlist}{\setlength{\itemsep}{0pt}\setlength{\parskip}{0pt}}
\newunicodechar{θ}{\ensuremath{\theta}}
\newunicodechar{γ}{\ensuremath{\gamma}}
\newunicodechar{π}{\ensuremath{\pi}}
\newunicodechar{ℓ}{\ensuremath{\ell}}
\newunicodechar{Δ}{\ensuremath{\Delta}}
\newunicodechar{Ω}{\ensuremath{\Omega}}
\newunicodechar{τ}{\ensuremath{\tau}}
\newunicodechar{α}{\ensuremath{\alpha}}
\newunicodechar{σ}{\ensuremath{\sigma}}
\newunicodechar{β}{\ensuremath{\beta}}
\newunicodechar{λ}{\ensuremath{\lambda}}

\title{Continued Training of Continuous Language Flows with Token Identity Squared Supervision}
\author[1,2]{Jiawei Jiang}
\author[1,2]{Zhaoxiang Zang\thanks{Corresponding author: \texttt{zhaoxiang.zang@kust.edu.cn}}}
\author[1,2]{Zhengtao Yu}
\affil[1]{Faculty of Information Engineering and Automation, Kunming University of Science and Technology, Kunming 650504, China}
\affil[2]{Key Laboratory of Artificial Intelligence in Yunnan Province, Kunming University of Science and Technology, Kunming 650504, China}
\date{}

\begin{document}
\maketitle 
\begin{abstract}
Diffusion and flow models have recently been extended to language modeling, generating text through continuous denoising in token embedding spaces. Pretrained models provide token representations and contextual prediction capabilities, but improving generation with limited additional training requires an appropriate connection between supervision and continuous generation. When continuous predictions are formed by weighting token embeddings with predicted probabilities, different probability distributions can yield the same vector and are therefore indistinguishable under a continuous reconstruction objective based on mean squared error (MSE). We propose Token Identity Brier Flow (TIBF) to apply squared supervision directly to token identities. TIBF fixes token embeddings, combines network scores with token-matching scores under a Gaussian observation model, supervises token probabilities using the Brier loss, and uses probability-weighted embeddings for continuous generation. We show that, for fixed token embeddings, the squared error in token posterior prediction controls the squared errors of the continuous denoising prediction and the flow velocity. A risk decomposition separates uncertainty inherent in the observation from posterior fitting error, motivating continued training that preserves pretrained embeddings and focuses on probability prediction. From common pretrained initializations and under matched continuation budgets, TIBF achieves mean Gen.PPL values of 34.87 on OWT and 80.08 on LM1B, outperforming continuous-MSE and CE controls. Further evaluations show improvements across multiple sampling budgets.
Code is available at \url{https://github.com/Jiangjiawei-kust/TIBF}.
\end{abstract}

\section{Introduction}
Language models must capture dependencies among tokens and translate them into an executable generation process. Autoregressive models, including Transformer decoders \citep{vaswani2017attention}, provide an established approach by predicting successive tokens in a prescribed order. Text completion, local editing, and coordination across multiple positions also depend on generation order and whether intermediate predictions can be revised. Diffusion language models generate text through repeated denoising or masked prediction, using context to coordinate updates across positions \citep{li2022diffusionlm,sahoo2024mdlm}.

Diffusion models \citep{sohldickstein2015diffusion,song2019scorematching,ho2020ddpm} and continuous-flow approaches such as flow matching (FM) and rectified flow \citep{lipman2022flowmatching,liu2022rectifiedflow} have achieved substantial progress on continuous data, including images. Diffusion training progressively corrupts data and learns a reverse denoising process; sampling reconstructs data from noise. Flow matching regresses velocity fields along probability paths to learn continuous transport from a noise distribution to the data distribution. Score-based formulations also connect reverse stochastic differential equations with corresponding probability-flow ordinary differential equations (ODEs) \citep{song2021sde}. These approaches provide a foundation for continuous language modeling.

Extending these approaches to text requires addressing token discreteness. Discrete diffusion perturbs tokens directly and learns to recover them through structured state transitions, discrete probability-ratio estimation, or masked prediction \citep{austin2021d3pm,lou2024sedd,sahoo2024mdlm}. Embedding-space methods instead represent tokens as continuous vectors and apply Gaussian corruption and denoising \citep{li2022diffusionlm,gulrajani2023plaid}. Continuous intermediate states permit iterative refinement. How continuous states represent token information and how predictions recover discrete text are therefore central to connecting representations, objectives, and generation rules.

Research on continuous language models has progressed from generation mechanisms toward joint design of objectives, likelihood evaluation, and scaling. LangFlow \citep{chen2026langflow} interprets token cross-entropy (CE) training through Bregman flow matching, combining information-uniform scheduling, self-conditioning, and an ODE-based upper bound on negative log-likelihood to connect identity prediction, continuous generation, and probabilistic evaluation. RePlaid \citep{yang2026replaid} retains Plaid's negative evidence lower bound (NELBO) objective while aligning its backbone with modern discrete diffusion models to compare scaling under equal training compute. These developments represent token-CE and variational-likelihood approaches to continuous language modeling.

When continuous predictions are formed by weighting token embeddings with predicted probabilities, different distributions can produce the same vector and are thus indistinguishable under continuous MSE. If embeddings are learned, the reconstruction metric also changes during training. Although CE directly supervises token identity, prior experiments on continued training of continuous language flows found that CE yielded no consistent Gen.PPL improvement, whereas MSE produced larger generation gains in the settings examined \citep{li2026convergeflow}. We therefore investigate squared supervision of the full token probability vector while retaining continuous generation, and relate identity prediction error to continuous denoising and flow velocities.

We propose Token Identity Brier Flow (TIBF). TIBF evaluates predicted probabilities against true token identities with the Brier squared loss, keeping the supervision metric fixed as embeddings are updated \citep{brier1950verification,shao2024scoringrules}. The model combines network scores with token-matching scores under a Gaussian observation model to predict token probabilities and weights embeddings by those probabilities to produce continuous denoising predictions.

Our risk decomposition shows that joint embedding updates change both the identity uncertainty inherent in the observations and the network's posterior fitting error, and that these changes can offset one another. Prior studies find that associations between token embeddings and semantic or syntactic features emerge early and stabilize during pretraining, and that preserving pretrained embeddings improves performance in some language-modeling settings \citep{papadimitriou2025vocabulary,welch2020embeddings}. Building on these existing representations, TIBF freezes pretrained token embeddings by default, preserving the observation channel and focusing additional training on posterior fitting. We denote the joint-update control as TIBF (trainable $E$) to evaluate the benefits of further representation adaptation. Because token information is recovered unevenly during denoising, TIBF also adopts information-uniform scheduling: denoising errors already computed during training are used to estimate information-change rates and dynamically allocate training across noise levels \citep{chen2026langflow,raya2026infonoise}.

On each of OWT and LM1B, all models start from the pretrained LangFlow checkpoint for that dataset. We compare TIBF, continuous MSE, and CE under matched token budgets for continued training and examine the effect of freezing or jointly updating the embeddings. We evaluate generation gains using Gen.PPL and supplement them with two dataset-PPL protocols for language modeling. Trajectory distances and decoding agreement characterize the correspondence between continuous states and fixed token embeddings at the endpoint. We also compare generation across sampling budgets; supplementary experiments examine quality--diversity trade-offs and zero-shot cross-corpus transfer.

Our contributions are threefold:

1. We propose TIBF for continued training of pretrained language flows, directly supervising full token probabilities with the Brier squared loss to address the inability of continuous MSE to distinguish probability predictions that map to the same vector. This squared identity metric remains fixed as embeddings are updated. The model combines network scores with token-matching scores under a Gaussian observation model to construct probabilities, then uses these same probabilities to weight embeddings for continuous generation.

2. We decompose risk into the minimum risk inherent in the observations and the network's posterior fitting error, showing that freezing embeddings focuses continued training on posterior fitting while joint updates can create tension between the two terms. With fixed embeddings, TIBF directly measures squared errors in the full token probability distribution, whereas flow-velocity error measures these probability deviations after the embedding map. TIBF thus directly supervises token identities while retaining a squared-error structure connected to continuous generation.

3. We demonstrate TIBF's generation improvements under matched pretrained initializations and continuation budgets. TIBF achieves the lowest mean Gen.PPL among the compared models on both OWT and LM1B, outperforming continuous MSE, CE, and TIBF (trainable $E$), with generation advantages across sampling budgets.

\section{Background and Problem Formulation}
\subsection{Flow Matching}
Flow matching (FM) learns a time-dependent velocity field that transforms a simple noise distribution into a data distribution \citep{lipman2022flowmatching}. The field specifies the direction and rate of state evolution. Let $x\sim p_{\mathrm{data}}$ be continuous data in $\mathbb R^d$, where $d$ is the data dimension and $p_{\mathrm{data}}$ is the continuous-data distribution. Let $\varepsilon\sim\mathcal N(0,I)$ be independent noise, with $I$ the identity matrix of matching dimension. For path time $t$ and state $z_t$, a common construction is a Gaussian probability path:
\begin{equation}
\label{eq:main:1}
    z_t=\alpha_t x+\sigma_t\varepsilon,\qquad t\in[0,1]
\end{equation}

The differentiable scalar coefficients $\alpha_t,\sigma_t$ scale data and noise, respectively, with $(\alpha_0,\sigma_0)=(0,1)$ and $(\alpha_1,\sigma_1)=(1,0)$. Thus, the distribution of $z_t$ interpolates from standard Gaussian noise to data. A model with parameters $\theta$ learns $v_\theta(z_t,t)$. Generation starts from noise and numerically solves the ODE $\mathrm dz_t/\mathrm dt=v_\theta(z_t,t)$.

The true marginal velocity is generally unavailable during training, but the conditional path velocity $\dot\alpha_t x+\dot\sigma_t\varepsilon$ is directly computable for sampled $x$ and $\varepsilon$. Conditional flow matching uses the squared regression objective
\begin{equation}
\label{eq:main:2}
\mathcal L_{\mathrm{CFM}}(\theta)
= \mathbb E_{t,x,\varepsilon}
\left[\left\|v_\theta(z_t,t)
-(\dot\alpha_t x+\dot\sigma_t\varepsilon)\right\|_2^2\right],
\qquad t\sim\operatorname{Unif}(0,1)
\end{equation}

Dots denote time derivatives, and $\|\cdot\|_2$ is the Euclidean norm. The expectation covers data $x$, noise $\varepsilon$, and an independent time $t$, with $\operatorname{Unif}(0,1)$ denoting the uniform distribution on $(0,1)$. The ideal predictor is the conditional mean of the path velocity, $u_t^*(z_t)=\mathbb E[\dot\alpha_t x+\dot\sigma_t\varepsilon\mid z_t,t]$. Under appropriate regularity conditions, this velocity field induces the marginal evolution of the specified probability path \citep{lipman2022flowmatching}.

For Gaussian paths, velocities can also be parameterized through clean-data predictions. Let $\widehat x_\theta(z_t,t)$ be a denoiser. For $0<t<1$ and $\sigma_t>0$,
\begin{equation}
\label{eq:main:3}
    v_\theta(z_t,t)
=\frac{\dot\sigma_t}{\sigma_t}z_t
+\left(\dot\alpha_t-\alpha_t\frac{\dot\sigma_t}{\sigma_t}\right)
\widehat x_\theta(z_t,t).
\end{equation}

Predicting the conditional mean of clean data therefore determines the generation velocity. With fixed representations and a fixed path, substituting this parameterization into conditional flow matching yields a noise-weighted continuous reconstruction MSE.

\subsection{Gaussian Paths over Token Embeddings}
Text consists of discrete tokens and cannot directly serve as continuous data on this Gaussian path. Embedding-space models first map tokens to vectors and then corrupt and denoise the resulting sequences \citep{li2022diffusionlm,gulrajani2023plaid}. Let $Y=(Y_1,\ldots,Y_L)\sim P_{\mathrm{data}}$ be a token sequence, where $P_{\mathrm{data}}$ is the distribution of discrete text, $Y_\ell\in[V]=\{1,\ldots,V\}$, $V$ is vocabulary size, $L$ is sequence length, and $\ell\in\{1,\ldots,L\}$ indexes positions. Let $E\in \mathbb R^{V\times d}$ be the token embedding matrix, where $d$ is the embedding dimension. Its $v$th row is $e_v^\top$, with $e_v\in \mathbb R^d$ a column vector. The superscript ${}^\top$ denotes transpose. The corresponding continuous sequence $X$ is
\begin{equation}
\label{eq:main:4}
X = X(Y) = 
\left[
\begin{array}{c}
e_{Y_1}^\top \\
\vdots \\
e_{Y_L}^\top
\end{array}
\right]
\in \mathbb R^{L\times d}.
\end{equation}

To express denoising difficulty directly, we follow LangFlow's negative log-SNR coordinate $\gamma=\log(\sigma^2/\alpha^2)$, where $\alpha,\sigma$ scale data and noise. Written as functions of $\gamma$, the noisy embedding state $Z_\gamma\in \mathbb R^{L\times d}$ follows a variance-preserving Gaussian path \citep{chen2026langflow}:
\begin{equation}
\label{eq:main:5}
Z_\gamma=\alpha_\gamma X+\sigma_\gamma\varepsilon,
\quad \alpha_\gamma^2=\operatorname{sigmoid}(-\gamma),
\quad \sigma_\gamma^2=\operatorname{sigmoid}(\gamma).
\end{equation}

Here $\operatorname{sigmoid}(a)=(1+e^{-a})^{-1}$, $\alpha_\gamma,\sigma_\gamma$ are positive square roots, and $\varepsilon\in \mathbb R^{L\times d}$ has independent standard Gaussian entries and is independent of $Y$. Variance preservation means $\alpha_\gamma^2+\sigma_\gamma^2=1$. Larger $\gamma$ corresponds to stronger noise, while smaller $\gamma$ corresponds to less corrupted token embeddings; the SNR is $\lambda_\gamma=e^{-\gamma}$. Generation proceeds in decreasing $\gamma$, with pure noise and clean data corresponding to $\gamma\to+\infty$ and $\gamma\to-\infty$, respectively.
\subsection{From Token Probabilities to Continuous Predictions}
Embedding-space generation connects token identity prediction with the continuous denoising predictions required by flow matching. Given the entire noisy sequence $Z_\gamma=Z$, the true token posterior and its probability-weighted embedding vector at position $\ell$ are
\begin{equation}
\label{eq:main:6}
q_{\ell,v}(Z,\gamma)=\Pr(Y_\ell=v\mid Z_\gamma=Z,\gamma),
\qquad
m_\ell^*(Z,\gamma)=E^\top q_\ell(Z,\gamma).
\end{equation}

Here $q_\ell=(q_{\ell,1},\ldots,q_{\ell,V})^\top$ is a $V$-dimensional posterior probability vector, and $m_\ell^*\in \mathbb R^d$ is the ideal continuous prediction. The probability simplex $\Delta_V=\{p\in \mathbb R_{\ge0}^V:\mathbf1_V^\top p=1\}$ contains nonnegative vectors summing to one, so $q_\ell\in\Delta_V$, where $\mathbf1_V$ is the all-ones column vector. Arranging positionwise probabilities as rows gives
\begin{equation}
\label{eq:main:7}
Q(Z,\gamma)=\big(q_{\ell,v}(Z,\gamma)\big)_{\ell,v}
\in \mathbb R^{L\times V}.
\end{equation}

The conditional mean of the clean embedding sequence is therefore
\begin{equation}
\label{eq:main:8}
m^*(Z,\gamma)=Q(Z,\gamma)E
= \mathbb E[X\mid Z_\gamma=Z,\gamma]
\in \mathbb R^{L\times d}.
\end{equation}

This conditional mean is the optimal denoising prediction under continuous squared error. The model approximates the true posterior with predicted probabilities $p_\ell\in\Delta_V$, assembled into $P\in \mathbb R^{L\times V}$. It then weights candidate embeddings by these probabilities to form $M=PE$ and determine the velocity field \citep{gulrajani2023plaid,chen2026langflow,li2026convergeflow}. Predictions at every position condition on the entire noisy sequence; positionwise outputs do not imply token independence.
\subsection{Supervision Objectives and the Problem Addressed}
The same continuous-generation interface supports different supervision targets. At one position, omit the position index and write $y=Y_\ell$ for the true token and $p=p_\ell$ for the predicted probability vector, whose component $p_y$ is the probability assigned to $y$. Continuous reconstruction MSE evaluates the probability-weighted embedding against the true token embedding \citep{li2026convergeflow}; CE evaluates identity prediction using the negative log probability $-\log p_y$ of the true token \citep{chen2026langflow}. Let $\delta_y\in \mathbb R^V$ be the one-hot identity label, with component $y$ equal to one. The continuous reconstruction error is
\begin{equation}
\label{eq:main:9}
    \frac1d\|E^\top p-e_y\|^2
=(p-\delta_y)^\top G_E(p-\delta_y),
\qquad G_E=\frac1dEE^\top.
\end{equation}

The matrix $G_E\in \mathbb R^{V\times V}$ is the probability-error metric induced by token embeddings. Thus, the probability-space metric of continuous MSE depends on $E$. In the usual setting $V>d+1$, distinct probability distributions can yield the same vector $E^\top p$; if $E$ is learned, this metric changes as well. In the extreme case where all token embeddings equal $c\in \mathbb R^d$, every probability prediction satisfies
\begin{equation}
\label{eq:main:10}
    E^\top(p-\delta_y)=c\,\mathbf1_V^\top(p-\delta_y)=0.
\end{equation}

CE directly supervises token identity, has the true posterior as its conditional optimum, and admits a flow-matching interpretation through Bregman divergences \citep{chen2026langflow}. We further investigate direct squared-loss supervision of token probabilities, allowing token probability prediction and continuous denoising prediction to be analyzed within a connected squared-error framework.

The Brier score evaluates the squared distance between a probability vector and a one-hot label \citep{brier1950verification,shao2024scoringrules}. Its metric is defined directly over token identities and does not change with the embeddings.

Figure~1 illustrates fixed equal-norm embeddings: two different probability predictions yield the same continuous vector and hence the same reconstruction MSE, while identity squared scoring distinguishes their allocations of probability to the true token.

\begin{figure}
    \centering
    \includegraphics[width=\linewidth]{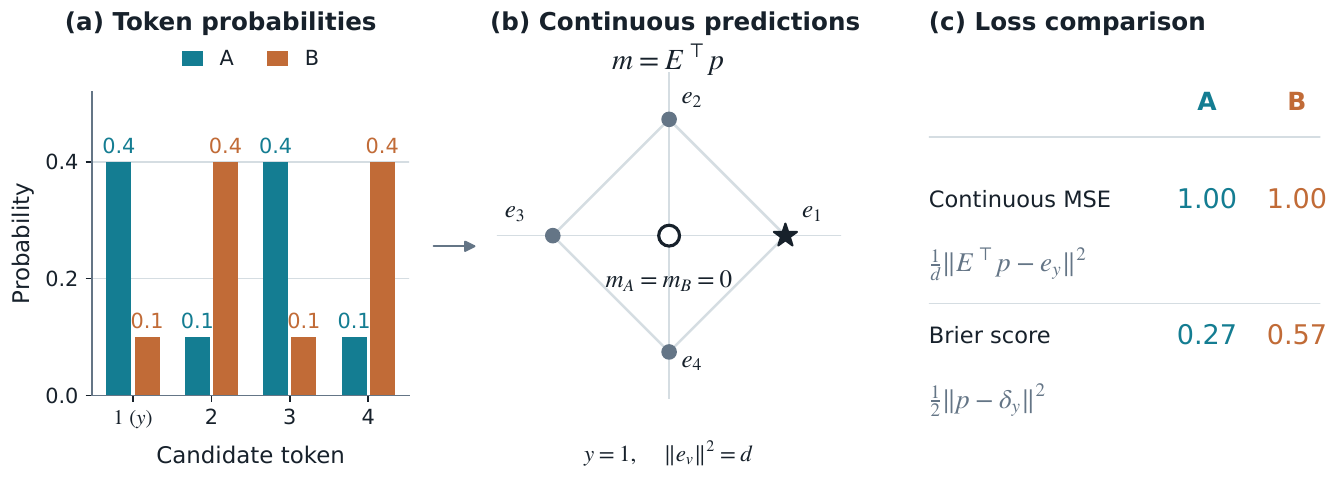}
    \caption{Continuous reconstruction versus token identity supervision. Probability predictions A and B map through the same embedding matrix to $m_A=m_B=0$. For the true token $y=1$, both have normalized continuous MSE 1, but half-Brier losses of 0.27 and 0.57, respectively. Embeddings satisfy $\|e_v\|^2=d$; geometric positions are shown as $e_v/\sqrt d$.}
    \label{fig1}
\end{figure}

\section{Method}

TIBF supervises token probabilities with the Brier squared loss and drives continuous generation through probability-weighted embeddings. We introduce the predictor and training objective, then use risk decomposition to analyze the role of frozen embeddings and the connection to flow-velocity error.

\subsection{Token Probability Prediction and Continuous Readout}
TIBF predicts token probabilities from noisy sequences and weights embeddings by those probabilities to obtain continuous denoising predictions. We retain the sequence length $L$, vocabulary size $V$, and embedding dimension $d$ from Section~2. For tokenwise probabilities, losses, and conditional risks, we fix one position and omit its index; predictions still condition on the entire noisy sequence.

\textbf{Embedding matrix and indexing.} Let $A\in \mathbb R^{V\times d}$ be the raw embedding parameters, with $a_v$ denoting the transpose of its $v$th row. The effective embedding is
\begin{equation}
\label{eq:main:11}
    e_v=\operatorname{RowNorm}(a_v)=\sqrt d\,
\frac{a_v}{\max(\|a_v\|,\varepsilon_{\mathrm{norm}})},
\qquad \varepsilon_{\mathrm{norm}}=10^{-12}.
\end{equation}

Here $a_v,e_v\in \mathbb R^d$, RowNorm denotes row normalization, and $\varepsilon_{\mathrm{norm}}$ prevents division by zero. If $\|a_v\|>\varepsilon_{\mathrm{norm}}$, each token embedding satisfies $\|e_v\|^2=d$. Our primary setting freezes pretrained embedding parameters $A$; the joint-update control updates $A$ together with the prediction network.

Let $z\in \mathbb R^d$ denote the noisy vector at the current position, the transpose of the corresponding row of $Z_\gamma$; $v$ indexes candidate tokens.

\textbf{Candidate scores from Gaussian observations.} By Eq.~\eqref{eq:main:5}, if the true token is $v$, its local observation follows
\begin{equation}
\label{eq:main:12}
z\mid(y=v,\gamma)
\sim\mathcal N(\alpha_\gamma e_v,\sigma_\gamma^2I_d).
\end{equation}

Here $I_d$ is the $d$-dimensional identity matrix. The Gaussian density is the product of the normalization factor $(2\pi\sigma_\gamma^2)^{-d/2}$ and the exponential term below. Since this factor is independent of candidate $v$, we use the exponential term to score candidate matches:
\begin{equation}
\label{eq:main:13}
K_v=
\exp\!\left[-\frac{\|z-\alpha_\gamma e_v\|^2}{2\sigma_\gamma^2}\right].
\end{equation}

The kernel $K_v$ measures how well the noisy vector $z$ matches candidate embedding $e_v$. To add it to network scores, we take its logarithm and expand:
\begin{equation}
\label{eq:main:14}
\log K_v
=-\frac{\|z\|^2}{2\sigma_\gamma^2}
+\frac{\alpha_\gamma}{\sigma_\gamma^2}z^\top e_v
-\frac{\alpha_\gamma^2\|e_v\|^2}{2\sigma_\gamma^2}.
\end{equation}

The first term is independent of candidate $v$. With equal-norm embeddings, the third term is also shared by all candidates, so both cancel during probability normalization. We therefore use the middle inner-product term as the Gaussian matching score, with coefficient $\kappa_\gamma=\alpha_\gamma/\sigma_\gamma^2$.

\textbf{Combining observations and context.} Let $\theta$ denote the prediction-network parameters, excluding embedding parameters $A$ and schedule parameters. The network reads the full noisy sequence and outputs $f_\theta(Z_\gamma,\gamma)\in \mathbb R^{L\times V}$; its base score for candidate $v$ at the current position is denoted by $f_v$. Network scores incorporate sequence context, whereas Gaussian matching scores evaluate the local noisy observation against candidate embeddings. We add these scores and normalize over the vocabulary:
\begin{align}
\label{eq:main:15}
\begin{aligned}
s_v&=f_v+\kappa_\gamma z^\top e_v,\\
p_v&=\frac{\exp(s_v)}
{\displaystyle\sum_{v'=1}^V\exp(s_{v'})}.
\end{aligned}
\end{align}

Here $s_v$ is the total score and $p_v$ is the candidate probability. Writing $p=(p_1,\ldots,p_V)^\top$, we have $p\in\Delta_V$.

\textbf{Equivalent Gaussian-kernel form.}

Define positive base weights $\omega_v=\exp(f_v)/\sum_{v'=1}^V\exp(f_{v'})$. Under equal-norm embeddings, Eq.~\eqref{eq:main:15} is equivalent to

\begin{equation}
\label{eq:main:16}
    p_v=
\frac{\omega_vK_v}
{\displaystyle\sum_{v'=1}^V\omega_{v'}K_{v'}}.
\end{equation}

Equation~\eqref{eq:main:16} shows that token probabilities are obtained by normalizing the products of network base weights and Gaussian kernels; the corresponding posterior factorization is derived in Appendix~\ref{sec:app:A.5}.

\textbf{Continuous readout.} Weighting embeddings by the local probabilities gives the continuous denoising prediction
\begin{equation}
\label{eq:main:17}
    m=E^\top p=\sum_{v=1}^Vp_ve_v\in \mathbb R^d.
\end{equation}

Because probabilities are nonnegative and sum to one, $m$ lies in $\operatorname{conv}\{e_1,\ldots,e_V\}$, the convex hull of the embeddings. This holds for any probability prediction and for both frozen and jointly updated $E$. Restoring position indices, arrange $p_\ell^\top$ as the rows of $P\in \mathbb R^{L\times V}$. The sequence-level prediction is $M=PE\in \mathbb R^{L\times d}$, with $m_\ell=E^\top p_\ell$ denoting the transpose of row $\ell$.

\textbf{From continuous predictions to generation velocities.} Equation~\eqref{eq:main:3} expresses the Gaussian-path velocity in terms of the current state and a clean-data prediction. Differentiating the coefficients in Eq.~\eqref{eq:main:5} with respect to $\gamma$ gives $\alpha_\gamma'=-\alpha_\gamma\sigma_\gamma^2/2$ and $\sigma_\gamma'=\sigma_\gamma\alpha_\gamma^2/2$, where primes denote these derivatives. Reparameterizing the path by $\gamma$ and substituting the continuous readout $M$ for the clean-data prediction in Eq.~\eqref{eq:main:3} yields

\begin{equation}
\label{eq:main:18}
    \begin{aligned}
v_\theta(Z_\gamma,\gamma)
&=\frac{\sigma_\gamma'}{\sigma_\gamma}Z_\gamma
+\left(\alpha_\gamma'-\alpha_\gamma\frac{\sigma_\gamma'}{\sigma_\gamma}\right)M\\
&=\frac12\left(\alpha_\gamma^2Z_\gamma-\alpha_\gamma M\right).
\end{aligned}
\end{equation}

Generation integrates this velocity field in the direction of decreasing $\gamma$, progressively denoising the state.
\subsection{Brier Supervision of Token Identity}
The continuous readout weights token embeddings by predicted probabilities. For training, TIBF directly compares the predicted probability vector $p$ with the true token's one-hot label $\delta_y$ using the half-Brier loss
\begin{equation}
\label{eq:main:19}
    \mathcal L(p,y)=\frac12\|p-\delta_y\|^2
=\frac12\left(1-2p_y+\sum_vp_v^2\right).
\end{equation}

Here $p_y$ is the probability assigned to the true token. The loss directly measures the squared distance between the full token probability vector and the true identity label.

To train across noise levels, independently draw $\gamma$ from distribution $\pi$ and construct noisy sequences using Eq.~\eqref{eq:main:5}. With positive noise-dependent weight $w(\gamma)>0$, the training risk is

\begin{equation}
\label{eq:main:20}
    \mathcal R(\theta,E)= \mathbb E\left[
\frac{w(\gamma)}L\sum_{\ell=1}^L\mathcal L(p_\ell,Y_\ell)\right].
\end{equation}
\subsection{Identity Risk Decomposition}
To analyze the difference between freezing and jointly updating token embeddings, we decompose the training risk into identity uncertainty that cannot be resolved from the observations and the prediction network's error in fitting the true posterior. We begin with the expected loss conditional on the observation. Retaining single-token notation, we define the true token posterior given the noisy sequence and noise level as
\begin{equation}
\label{eq:main:21}
q_v=\Pr(y=v\mid Z_\gamma,\gamma),\qquad
q=(q_1,\ldots,q_V)^\top\in\Delta_V.
\end{equation}

The true posterior is also the conditional mean of the one-hot label:
\begin{equation}
\label{eq:main:22}
\mathbb E[\delta_y\mid Z_\gamma,\gamma]=q.
\end{equation}

For a fixed observation and prediction $p$, expanding the Brier loss and taking the conditional expectation over the true token using Eq.~\eqref{eq:main:22} gives
\begin{equation}
\label{eq:main:23}
\mathbb E[\mathcal L(p,y)\mid Z_\gamma,\gamma]
=\frac12\|p-q\|^2
+\frac12\left(1-\|q\|^2\right).
\end{equation}

The first term is half the squared error between predicted probabilities and the true posterior; the second is the minimum expected loss due to residual identity uncertainty in the observation. Thus, the conditional expected loss is minimized at $p=q$, where the first term is zero. Training still uses one-hot labels: Eq.~\eqref{eq:main:23} decomposes their expected loss and does not require computing the posterior. See Appendix~\ref{sec:app:A.1} for the proof.

The true posterior is determined by the data distribution, $E$, and corruption process, independently of network parameters $\theta$. Taking the weighted expectation of the second term as in the training risk gives the minimum risk of the observation channel
\begin{equation}
\label{eq:main:24}
\mathcal R^*(E)
:= \mathbb E\left[\frac{w(\gamma)}{2L}
\sum_\ell(1-\|q_\ell\|^2)\right].
\end{equation}

The first term of Eq.~\eqref{eq:main:23} yields the model's posterior fitting error:
\begin{equation}
\label{eq:main:25}
    \mathcal E(\theta,E)
= \mathbb E\left[\frac{w(\gamma)}{2L}
\sum_\ell\|p_\ell-q_\ell\|^2\right].
\end{equation}

\textbf{Proposition 1: Identity risk decomposition.} The population risk satisfies
\begin{equation}
\label{eq:main:26}
    \boxed{\quad
\mathcal R(\theta,E)=\mathcal R^*(E)+\mathcal E(\theta,E).
\quad}
\end{equation}

With the data distribution, noise distribution, and weighting fixed, freezing $E$ keeps $\mathcal R^*(E)$ constant. Hence,
\begin{equation}
\label{eq:main:27}
    \mathcal R(\theta,E)-\mathcal R^*(E)=\mathcal E(\theta,E).
\end{equation}

With frozen $E$, reducing training risk is equivalent to reducing squared error between predicted probabilities and the true posterior.

\textbf{Frozen versus joint updates.} Joint updates to $E$ change both minimum risk and posterior fitting error. These terms can offset one another, so the observations may become less informative about token identity even as total risk decreases. Given existing pretrained representations, freezing $E$ preserves the observation channel and focuses additional training on posterior fitting. Appendix~\ref{sec:app:B.2} analyzes the risk comparison between the two settings when the embeddings have little remaining room for improvement. Section~4.3 compares their empirical performance after continued training.

\textbf{Risk decomposition and flow velocity.} Let $r=p-q$. With token embeddings $E$ and the noise level fixed, Eq.~\eqref{eq:main:18} gives
\begin{equation}
\|v_\theta-v^*\|^2
=\frac{\alpha_\gamma^2}{4}\|E^\top r\|^2,
\label{eq:main:added1}
\end{equation}
where $v^*$ is the ideal velocity constructed from the posterior-weighted embedding vector $E^\top q$.

Ignoring the fixed dimension-normalization factor, the posterior-fitting term for continuous MSE is
\begin{equation}
\|E^\top p-E^\top q\|^2=\|E^\top r\|^2,
\label{eq:main:added2}
\end{equation}
which is proportional to the squared velocity error at a fixed noise level. It measures deviations in the probability-weighted embedding vector and cannot distinguish probability predictions that yield the same continuous readout.

The posterior-fitting term for TIBF is
\begin{equation}
\frac12\|p-q\|^2=\frac12\|r\|^2.
\label{eq:main:added3}
\end{equation}
It directly measures squared deviations of token probabilities from the true posterior, whereas denoising prediction and velocity errors measure these deviations after applying the embedding map. TIBF therefore constrains probability deviations that the continuous readout cannot distinguish while retaining a squared-error form connected to continuous denoising prediction.

CE also has the true posterior as its optimal prediction, with the posterior-fitting term
\begin{equation}
\mathrm{KL}(q\|p)
=\sum_{v:q_v>0}q_v\log\frac{q_v}{p_v}.
\label{eq:main:added4}
\end{equation}
CE measures posterior fitting through a logarithmic score. TIBF instead directly supervises token probabilities with squared error, allowing token probability prediction, continuous denoising prediction, and flow velocity to be analyzed within a connected squared-error framework. Appendix~\ref{sec:app:A.1}--\ref{sec:app:A.2} provides the risk decompositions for all three objectives and further relations between probability errors and continuous errors.

\section{Experiments}
On OWT and LM1B, we compare continuous MSE, CE, and TIBF from common pretrained initializations under equal continuation budgets, examining generation, language modeling, and the effects of frozen versus jointly updated embeddings. Trajectory diagnostics further examine the low-noise correspondence among the continuous state, continuous prediction, and token probabilities of actual predictors.

\subsection{Data, Models, and Training Setup}

\textbf{Data and shared initialization.} OpenWebText (OWT) \citep{gokaslan2019openwebtext} uses the 50,257-token GPT-2 vocabulary and sequence length 1,024; the One Billion Word Benchmark (LM1B) \citep{chelba2013billionword} uses bert-base-uncased and length 128. Within each dataset, all groups inherit the same LangFlow backbone, vocabulary head, $E$, and schedule parameters, with fresh optimizers and exponential moving average (EMA) states. Data splits for both datasets are specified in Appendix~\ref{sec:app:D.data}.

\textbf{Models.} All continuation models retain LangFlow's DiT architecture with 12 layers, token embedding and hidden dimensions both set to 768, 12 attention heads, rotary positional embeddings, and dropout 0.1, together with the same vocabulary head, self-conditioning module, and Gaussian observation branch. Including token embeddings, the OWT and LM1B models contain approximately 171M and 140M parameters, respectively. Frozen settings preserve pretrained $E$; joint settings update it together with the network.

\textbf{Noise scheduling.} TIBF follows information-uniform scheduling: it estimates information-change rates from current denoising errors and fits a Gumbel distribution online to allocate more training samples to noise regions where identity information changes rapidly \citep{chen2026langflow,raya2026infonoise}. The derivation and implementation appear in Appendices~\ref{sec:app:A.6} and \ref{sec:app:D.3}.

\textbf{Self-conditioning.} All groups retain LangFlow's self-conditioning setup \citep{chen2026langflow}. During training, with probability 0.25, we compute a preliminary prediction at the same noisy state without gradients and supply its continuous denoising readout as an auxiliary input. Generation uses empty self-conditioning at the first point and reuses the preceding node's denoising prediction thereafter. Construction and stop-gradient details appear in Appendix~\ref{sec:app:D.7}.

\textbf{Optimization and budget.} All five methods are independently continued for 100,000 updates per run. Effective batch sizes are 6 on OWT and 48 on LM1B, both yielding approximately 614.4M additional tokens. We use AdamW with learning rate $3\times10^{-4}$, a 2,500-step warmup followed by a constant rate, EMA decay 0.9999, and BF16 mixed precision. Evaluation uses the final EMA. Further settings appear in Appendix~\ref{sec:app:D.2}.

\textbf{Controls.} Both OWT and LM1B include five continuation models: ConvergeFlow freezes pretrained embeddings and trains with continuous MSE; LangFlow uses CE with either frozen or trainable embeddings; TIBF uses the half-Brier loss, with TIBF (trainable $E$) as its joint-update control. The controls retain the training objectives of the original methods and are compared from common pretrained initializations under equal additional-token budgets. Training and scheduling details appear in Appendix~\ref{sec:app:D}.

\subsection{Evaluation Metrics and Comparison Protocol}

\textbf{Generation quality.} We compute generative perplexity (Gen.PPL) with GPT-2 large, measuring the predictability of generated text under a fixed external language model; lower values indicate a higher average log probability assigned by the scoring model \citep{radford2019gpt2}. Each setting generates 1,024 sequences of length 1,024 for OWT or 128 for LM1B. Outputs from both datasets are decoded to text and retokenized with the scorer's tokenizer. Scoring details are given in Appendix~\ref{sec:app:E.1}.

\textbf{Language modeling.} On the same set of 1,024 held-out sequences within each dataset, we report two probability-flow ODE protocols for dataset PPL. The attached protocol uses the full derivative through same-state self-conditioning to estimate an upper bound on negative log-likelihood. The detached protocol holds the self-conditioning input fixed when computing divergence and is reported as a separate evaluation score. Both protocols are reported in Tables~\ref{tab1} and \ref{tab:lm1b_comparison}; the derivation and numerical settings appear in Appendix~\ref{sec:app:E}.

\textbf{Diversity.} Token entropy and repeated 4-gram fraction supplement generation-quality evaluation. Token entropy measures within-sequence dispersion of token usage; repeated 4-gram fraction measures repetition of consecutive four-token combinations. Both are computed per sequence and then averaged, and are interpreted jointly with Gen.PPL. Definitions are given in Appendix~\ref{sec:app:F.metrics}, and quality--diversity comparisons in Appendix~\ref{sec:app:F.quality}.

\textbf{Sampling and statistics.} All methods use the same Euler rule; NFE counts actual network evaluations, including final decoding. The primary evaluation uses temperature 1. Primary evaluations, sampling-budget sweeps, temperature studies, and cross-corpus transfer experiments report means and sample standard deviations across three training seeds. Paired confidence intervals are computed for differences in mean Gen.PPL. Details appear in Appendices~\ref{sec:app:D.4}--\ref{sec:app:D.5} and \ref{sec:app:E.1}.

\subsection{Continued Training from a Common Pretrained Initialization}
\subsubsection{Generation and Modeling under Equal Continuation Budgets}
The primary evaluations use NFE256 for OWT and NFE128 for LM1B; results are reported in Tables~\ref{tab1} and \ref{tab:lm1b_comparison}.
\begin{table}[htbp]
    \centering
    \caption{OWT continuation from a common pretrained initialization. Mean $\pm$ sample SD.}
    \label{tab1}
    \begin{adjustbox}{max width=\linewidth}
\begin{tabular}{lccc}
\toprule
Model & Gen.PPL & \shortstack{Dataset PPL\\attached} & \shortstack{Dataset PPL\\detached} \\ \midrule
ConvergeFlow & $43.71\pm1.23$ & $40.41\pm0.17$ & $27.28\pm0.14$ \\
LangFlow (trainable $E$) & $53.03\pm0.84$ & $39.17\pm0.05$ & $29.14\pm0.16$ \\
LangFlow (frozen $E$) & $53.59\pm0.46$ & $\boldsymbol{38.39\pm0.22}$ & $28.29\pm0.13$ \\
TIBF (trainable $E$) & $35.84\pm2.07$ & $44.76\pm0.04$ & $30.59\pm0.17$ \\
TIBF & $\boldsymbol{34.87\pm0.98}$ & $41.62\pm0.10$ & $\boldsymbol{27.22\pm0.15}$ \\
\bottomrule\end{tabular}
\end{adjustbox}
\end{table}
\begin{table}[htbp]
    \centering
    \caption{LM1B continuation from a common pretrained initialization. Mean $\pm$ sample SD.}
    \label{tab:lm1b_comparison}
    \begin{adjustbox}{max width=\linewidth}
\begin{tabular}{lccc}
\toprule
Model & Gen.PPL & \shortstack{Dataset PPL\\attached} & \shortstack{Dataset PPL\\detached} \\ \midrule
ConvergeFlow & $83.63\pm1.01$ & $54.38\pm0.13$ & $35.36\pm0.02$ \\
LangFlow (trainable $E$) & $92.21\pm1.84$ & $46.85\pm0.02$ & $34.10\pm0.06$ \\
LangFlow (frozen $E$) & $89.87\pm0.60$ & $\boldsymbol{46.36\pm0.06}$ & $\boldsymbol{33.49\pm0.04}$ \\
TIBF (trainable $E$) & $85.79\pm1.03$ & $56.07\pm2.99$ & $37.53\pm1.43$ \\
TIBF & $\boldsymbol{80.08\pm0.63}$ & $55.45\pm0.20$ & $35.36\pm0.01$ \\
\bottomrule\end{tabular}
\end{adjustbox}
\end{table}

TIBF achieves Gen.PPL of $34.87\pm0.98$ on OWT and $80.08\pm0.63$ on LM1B, the lowest means among the compared models and reductions of 20.21\% and 4.24\% relative to continuous MSE, respectively. Within TIBF, freezing also yields lower mean Gen.PPL than joint embedding updates on both corpora, supporting the preservation of pretrained embeddings under the current continuation budget.

Dataset PPL yields different comparisons: LangFlow (frozen $E$) achieves the lowest mean under attached evaluation on both datasets. Under detached evaluation, TIBF has the lowest mean on OWT and is close to continuous MSE on LM1B, where LangFlow (frozen $E$) achieves the lowest mean.
Quality--diversity comparisons, temperature sensitivity, paired statistics, and zero-shot transfer appear in Appendices~\ref{sec:app:F.quality}--\ref{sec:app:F.transfer}.

\subsubsection{Generation across Sampling Budgets}\label{sec:main:budget}
We evaluate the final checkpoints and vary only actual NFE: 64, 128, 256, 512, and 1,024 on OWT; 16, 32, 64, and 128 on LM1B. All other generation and scoring settings follow the primary protocol for each corpus. Tables~\ref{tab:owt_budget} and \ref{tab:lm1b_budget} report means and sample standard deviations.
\begin{table}[htbp]
    \centering
    \caption{Gen.PPL across sampling budgets for OWT continuation models \(\downarrow\).}
    \label{tab:owt_budget}
    \begin{adjustbox}{max width=\linewidth}
\begin{tabular}{lccccc} 
        \toprule
        Model & NFE=64 & NFE=128 & NFE=256 & NFE=512 & NFE=1024 \\
        \midrule
ConvergeFlow & $72.13\pm1.86$ & $54.24\pm1.07$ & $43.71\pm1.23$ & $36.65\pm1.52$ & $31.60\pm1.79$ \\
LangFlow (trainable $E$) & $84.04\pm0.92$ & $64.04\pm0.79$ & $53.03\pm0.84$ & $45.92\pm1.22$ & $40.89\pm1.19$ \\
LangFlow (frozen $E$) & $83.00\pm0.63$ & $64.14\pm0.54$ & $53.59\pm0.46$ & $46.53\pm0.13$ & $41.74\pm0.69$ \\
TIBF (trainable $E$) & $65.39\pm2.52$ & $47.15\pm2.10$ & $35.84\pm2.07$ & $27.94\pm1.81$ & $22.44\pm1.52$ \\
TIBF & $\boldsymbol{64.24\pm4.58}$ & $\boldsymbol{45.77\pm2.02}$ & $\boldsymbol{34.87\pm0.98}$ & $\boldsymbol{27.19\pm0.29}$ & $\boldsymbol{22.11\pm0.17}$ \\
\bottomrule
    \end{tabular}
\end{adjustbox}
\end{table}
\begin{table}[htbp]
    \centering
    \caption{Gen.PPL across sampling budgets for LM1B continuation models \(\downarrow\).}
    \label{tab:lm1b_budget}
    \begin{adjustbox}{max width=\linewidth}
\begin{tabular}{lcccc} 
        \toprule
        Model & NFE=16 & NFE=32 & NFE=64 & NFE=128 \\
        \midrule
ConvergeFlow & $163.42\pm1.32$ & $115.14\pm0.69$ & $94.21\pm0.70$ & $83.63\pm1.01$ \\
LangFlow (trainable $E$) & $174.08\pm2.07$ & $124.91\pm2.34$ & $102.44\pm2.45$ & $92.21\pm1.84$ \\
LangFlow (frozen $E$) & $169.85\pm0.48$ & $120.58\pm0.93$ & $100.06\pm0.74$ & $89.87\pm0.60$ \\
TIBF (trainable $E$) & $170.55\pm3.04$ & $118.88\pm0.94$ & $96.43\pm0.65$ & $85.79\pm1.03$ \\
TIBF & $\boldsymbol{156.40\pm1.01}$ & $\boldsymbol{110.96\pm0.56}$ & $\boldsymbol{90.23\pm0.87}$ & $\boldsymbol{80.08\pm0.63}$ \\
\bottomrule
    \end{tabular}
\end{adjustbox}
\end{table}

TIBF achieves the lowest mean Gen.PPL at all five sampling budgets on OWT and all four on LM1B, showing that its generation gains over continuous MSE and CE continuation extend across sampling budgets. Diversity across sampling budgets is reported in Appendix~\ref{sec:app:F.budget}.

\subsubsection{Endpoint Convergence Diagnostics with Fixed Pretrained Embeddings}\label{sec:main:trajectory}
Using the final EMA of TIBF trained on OWT with seed 17, we examine whether continuous states approach token embeddings during denoising and whether state nearest neighbors, prediction nearest neighbors, and probability argmax identify the same token. The ideal endpoint distribution and sufficient conditions for local convergence are established in Appendix~\ref{sec:app:C}.

\begin{figure}
    \centering
    \includegraphics[width=\linewidth]{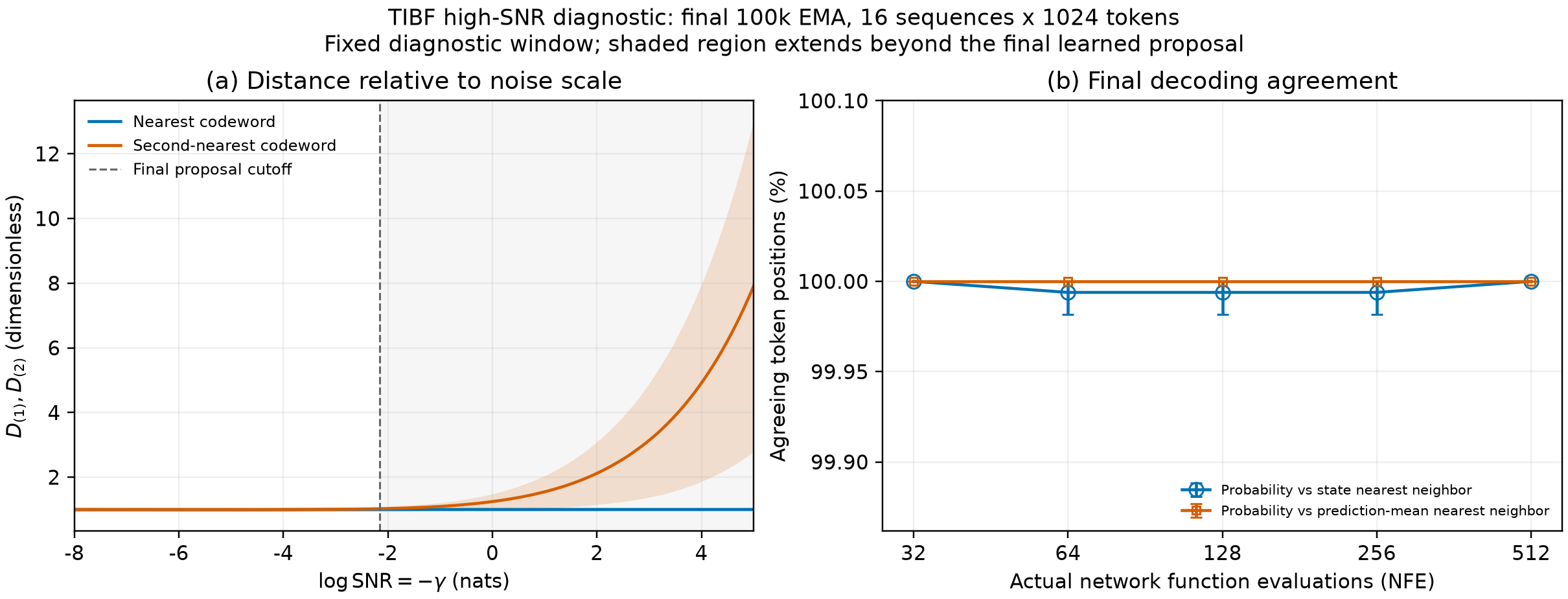}
    \caption{Empirical trajectory convergence of TIBF. (a) Noise-normalized distances to the nearest and second-nearest embeddings. Solid curves show midpoints of the 1\%--99\% quantile intervals; shaded regions show the intervals. (b) Endpoint agreement of probability argmax with state and continuous-prediction nearest neighbors. Error bars are 95\% sequence-bootstrap intervals. The dashed line marks the final schedule's low-noise boundary.}
    \label{fig2}
\end{figure}

\textbf{Trajectory proximity and candidate separation.} For a state $z$ at one position, define the noise-normalized distance to candidate embedding $e_j$ as $D_j=\|z-\alpha e_j\|/(\sigma\sqrt d)$ \citep{li2026convergeflow}, where $\alpha,\sigma$ are path coefficients and $d$ is the embedding dimension. Denote the nearest and second-nearest distances by $D_{(1)},D_{(2)}$; Figure~\ref{fig2}(a) shows their evolution during denoising. The diagnostic uses $\mathrm{logSNR}=-\gamma$, ranging from $-8$ to 5, with a uniform grid in path time. The region beyond the final schedule's low-noise boundary is shaded gray. Implementation details appear in Appendix~\ref{sec:app:G.1}.

At low SNR, the normalized distances to the nearest and second-nearest candidates nearly coincide. As SNR increases, the nearest-candidate curve stays close to 1, indicating that the distance from the state to the nearest candidate center $\alpha e_j$ shrinks with the noise scale $\sigma\sqrt d$. The normalized distance to the second-nearest candidate rises from about 1 to 7.91, widening the gap in units of the noise scale.

\textbf{Endpoint decoding agreement.} With the same diagnostic endpoints and NFE ranging from 32 to 512, state nearest-neighbor decoding agrees with probability argmax at 99.9939\%--100\% of positions, while prediction nearest-neighbor decoding agrees at 100\%; see Figure~\ref{fig2}(b). Candidate separation and high decoding agreement support the correspondence among token probabilities, continuous predictions, and state endpoints analyzed in Appendix~\ref{sec:app:C.8}. Appendices~\ref{sec:app:G.2} and~\ref{sec:app:G.3} report distances to fixed targets under continued denoising and recovery after local perturbations.

\section{Related Work}
\subsection{Diffusion and Continuous Language Flows}
Diffusion language models enable parallel generation by repeatedly recovering text from corrupted states. Among discrete approaches, D3PM unifies corruption mechanisms through structured transition matrices \citep{austin2021d3pm}, SEDD learns probability ratios between discrete states through score entropy \citep{lou2024sedd}, and MDLM exploits absorbing states and Rao--Blackwellization to simplify training to weighted language modeling at different masking levels \citep{sahoo2024mdlm}. These methods design corruption and recovery directly in token space.

Continuous approaches embed tokens as vectors and generate through Gaussian denoising or flow matching. Diffusion-LM enables fine-grained control, including syntactic constraints, through gradient guidance on continuous intermediate states \citep{li2022diffusionlm}. Self-conditioned Embedding Diffusion feeds previous clean-representation predictions back as auxiliary inputs and uses stop-gradient preliminary predictions during training to simulate this feedback and improve embedding-space denoising \citep{strudel2022sed}. Plaid jointly trains embeddings and a generator with a negative evidence lower bound (NELBO), enabling explicit likelihood modeling with continuous diffusion \citep{gulrajani2023plaid}. RePlaid retains this likelihood objective, aligns modern backbones and training configurations, and reevaluates continuous diffusion through scaling experiments with matched compute \citep{yang2026replaid}.

JPEG-DLM and CALM learn compressed representations to reduce the positions processed during continuous generation and the sequential steps of autoregressive generation, respectively \citep{yuan2026jpegdlm,shao2025calm}.

Language flows further examine the interface between continuous prediction and discrete tokens. LangFlow gives CE training a flow-matching interpretation through Bregman divergences, forms continuous predictions using probability-weighted embeddings, and derives an ODE-based upper bound on negative log-likelihood \citep{chen2026langflow}. ELF primarily performs flow matching in continuous embedding space and maps endpoints to tokens with a shared-weight network, facilitating techniques such as classifier-free guidance \citep{hu2026elf}. FLM constructs continuous flows on one-hot representations using a CE objective compatible with the probability simplex; FMLM distills flow maps across time to reduce integration steps \citep{lee2026flowmap}. ConvergeFlow combines network base weights and Gaussian kernels into convex combinations of token embeddings and supervises continuous predictions with flow-matching-induced MSE \citep{li2026convergeflow}.

These approaches connect continuous states to discrete text through token-probability supervision, continuous reconstruction, or both. Continuous MSE evaluates probability-weighted embeddings, but different token probabilities can yield the same continuous prediction.

\subsection{Proper Scoring Rules and Generative Objectives}
Proper scoring rules are optimized in expectation by the true distribution; strict propriety requires this optimum to be unique. The Brier score measures squared distance between the full probability vector and the event label \citep{brier1950verification}. Gneiting and Raftery characterize proper scoring rules through convex functions, entropies, and Bregman divergences, establishing objectives with consistent probability optima beyond logarithmic scoring \citep{gneiting2007proper}. Br\"ocker generalizes classical score decompositions to proper scoring rules, separating uncertainty, resolution, and reliability to distinguish predictive information from fitting quality \citep{brocker2009decomposition}.

For language generation, Shao et al. use tokenwise conditional prediction to avoid the exponentially large sentence-level sample space, enabling nonlocal scores such as Brier and spherical scoring for autoregressive training \citep{shao2024scoringrules}. For continuous outputs, EAR trains visual autoregressive models with a sample-based energy score, avoiding explicit continuous-density computation \citep{shao2025ear}. Distributional Diffusion uses scoring rules to learn the clean-data distribution conditional on noisy observations, then constructs coarse-step reverse transitions through posterior sampling \citep{debortoli2025distributional}.

These studies provide a foundation for generative learning with nonlogarithmic proper scores. We apply identity squared scoring to continuous language flows and analyze how the embedding map connects probability error to continuous generation.

\subsection{Token Representations and Continued Training}
Continuous language models must account for representational distinguishability, denoising difficulty, and reliable decoding. In vector-quantized image modeling, Nguyen et al. analyze embedding collapse under end-to-end joint training and propose VQ-LCMD, which stabilizes representations through a joint variational objective, consistency-matching regularization, and corresponding noise-training strategies \citep{nguyen2024catdm}. RePlaid examines embedding geometry under likelihood training and uses auxiliary-CE ablations to relate objective changes to representation structure and modeling performance \citep{yang2026replaid}. These findings motivate evaluating embedding training together with the generation objective rather than using geometric dispersion alone as a quality criterion.

Continued training examines how existing representations can improve generation. \citet{papadimitriou2025vocabulary} track Pythia and OLMo checkpoints and find that correlations between embedding geometry and semantic or syntactic features stabilize early, while the vectors continue to change later in training. \citet{welch2020embeddings} find that pretraining and freezing input embeddings improves low-compute word-level language modeling, with gains primarily attributable to rare-word representations. Preserving consistency in the existing embedding space motivates their freezing strategy. These studies provide an empirical basis for continued training that builds on existing lexical structure. ConvergeFlow starts from a common LangFlow checkpoint, freezes pretrained embeddings, and studies generation gains through MSE, CE, and objective-switching experiments, finding that CE yielded no consistent Gen.PPL improvement, whereas MSE produced larger generation gains in its settings \citep{li2026convergeflow}.

\subsection{Information-Based Noise Scheduling}
Noise scheduling determines how training samples denoising tasks of different difficulty. Variational Diffusion Models exploit the invariance of the continuous-time variational bound to time parameterization at fixed endpoint SNRs, learning schedules by minimizing estimator variance \citep{kingma2021vdm}. RePlaid follows this likelihood-based approach and examines its connection to approximately uniform information progress \citep{yang2026replaid}.

Information-based approaches allocate training according to information change during denoising. LangFlow approximates token conditional entropy with CE, fits an information curve using a trainable Gumbel CDF, and uses its rate of change to allocate samples and generation nodes \citep{chen2026langflow}. InfoNoise uses the Gaussian-channel I--MMSE relation to estimate conditional-entropy change rates online from existing denoising errors and adjusts the noise distribution while retaining the main objective, loss weighting, and parameterization \citep{guo2005immse,raya2026infonoise}.

\subsection{Continuous Trajectories and Token Endpoints}
Flow matching connects tractable conditional-path training to target marginal distributions through conditional-mean velocity fields, providing a foundation for continuous language flows \citep{lipman2022flowmatching}. Wan et al. analyze ODE trajectories through denoiser geometry: early and intermediate trajectories are attracted to data means and local clusters, respectively, while terminal analysis establishes convergence results including low-dimensional data supports \citep{wan2025flowdynamics}. These results characterize distributional evolution and continuous-generation geometry; language generation additionally requires connecting endpoints to token identities.

For Gaussian-weighted embedding predictors, ConvergeFlow proves convergence in probability to token endpoints with prediction error under positive base weights, uniform trajectorywise log-Lipschitz regularity, and suitable grid conditions \citep{li2026convergeflow}.

\clearpage
\section{Conclusion}
We introduce TIBF for continued training of pretrained continuous language flows, addressing the inability of continuous MSE to distinguish different token probability distributions that map to the same embedding vector. TIBF directly supervises the full token distribution with the Brier squared loss and drives continuous generation through probability-weighted embedding vectors. Our risk decomposition shows that freezing pretrained embeddings focuses continued training on posterior fitting. With token embeddings fixed, we further bound continuous denoising and flow-velocity errors by the squared error in token posterior prediction. Endpoint analysis and trajectory diagnostics further support the connection between continuous generation and token decoding. With a common pretrained initialization and matched budgets for continued training, TIBF achieves the lowest mean Gen.PPL among the compared models on both OWT and LM1B, with generation gains across multiple sampling budgets and different quality--diversity trade-offs. This work provides a squared-risk perspective on the connection between token identity supervision and continuous generation, and an effective method for improving generation from existing language flows through continued training. Future work will examine its applicability to larger models and longer continuation budgets, as well as training strategies that balance generation quality and diversity.

\clearpage
\bibliographystyle{plainnat}
\bibliography{references}

\clearpage
\appendix
\setcounter{table}{0}
\setcounter{figure}{0}
\renewcommand{\thetable}{\Alph{section}.\arabic{table}}
\renewcommand{\thefigure}{\Alph{section}.\arabic{figure}}
\input{appendices}
\end{document}

%% file: appendices.tex
\newcounter{tibfappanchor}
\AtBeginEnvironment{equation}{\stepcounter{tibfappanchor}}
\renewcommand*{\theHequation}{tibfapp.\arabic{tibfappanchor}}
\counterwithin*{table}{section}
\counterwithin*{figure}{section}
\renewcommand*{\theHtable}{tibfapp.\arabic{section}.\arabic{table}}
\renewcommand*{\theHfigure}{tibfapp.\arabic{section}.\arabic{figure}}

\section{Theoretical Supplements for Identity Supervision and Noise Scheduling}\label{sec:app:A}

\subsection{Risk Decompositions for Brier, MSE, and CE}\label{sec:app:A.1}

Noise is sampled over a finite interval, the weight in Eq.~\eqref{eq:app:D.1} is bounded, and both half-Brier loss and posterior fitting error are bounded. The risks in Eqs.~\eqref{eq:main:24}--\eqref{eq:main:26} are therefore finite. Throughout this section, minimum risk means the smallest conditional expected loss over the full probability simplex.

Retain the posterior matrix $Q$ defined in the main text. Let \(\Omega\) denote prediction randomness independent of the data and the corruption process; it can be omitted for deterministic predictors. Given \((Z_\gamma,\gamma,\Omega)\), $p$ is fixed and the label posterior remains $q$. Since \(\|\delta_{Y_\ell}\|^2=1\) and \(\mathbb E[\delta_{Y_\ell}\mid Z_\gamma,\gamma,\Omega]=q_\ell\),

\begin{equation}
\mathbb E[\mathcal L(p_\ell,Y_\ell)\mid Z_\gamma,\gamma,\Omega]
=\frac12\|p_\ell\|^2-p_\ell^\top q_\ell+\frac12.
\tag{A.1}\label{eq:app:A.1}
\end{equation}

Completing the square yields Eq.~\eqref{eq:main:23}. Averaging over positions and taking weighted expectations over data, noise levels, Gaussian perturbations, and independent prediction randomness gives

\begin{equation}
\mathcal R(\theta,E)
=\mathbb E\left[\frac{w(\gamma)}{2L}\sum_\ell(1-\|q_\ell\|^2)\right]
+\mathbb E\left[\frac{w(\gamma)}{2L}\sum_\ell\|p_\ell-q_\ell\|^2\right].
\tag{A.2}\label{eq:app:A.2}
\end{equation}

The first term is independent of the predictor; the second vanishes at $P=Q$. They are therefore the minimum risk and posterior fitting error, respectively. The decomposition uses positionwise marginal posteriors while retaining dependencies across sequence positions.

Fix one position and use the main text's single-token notation $y$. Conditional on the observation and prediction randomness, $p$ is fixed and the true token has distribution $q$.

\textbf{Continuous MSE.} The per-token loss is $\|E^\top p-e_y\|^2$. Write the error as $(E^\top p-E^\top q)+(E^\top q-e_y)$. Since $\sum_vq_ve_v=E^\top q$, the cross term vanishes in conditional expectation, yielding
\begin{equation}
\sum_vq_v\|E^\top p-e_v\|^2
=\underbrace{\sum_vq_v\|e_v-E^\top q\|^2}_{\text{minimum conditional risk}}
+\underbrace{\|E^\top(p-q)\|^2}_{\text{posterior-fitting term}}.
\tag{A.3}
\label{eq:app:A.added1}
\end{equation}
The first term is independent of $p$, while the second is nonnegative and vanishes exactly when $E^\top p=E^\top q$. Thus continuous MSE attains its minimum when the readouts agree, without requiring $p=q$. In the experiments, continuous MSE is averaged over embedding coordinates, dividing both the loss and each term of its decomposition by $d$.

\textbf{CE.} The per-token loss is $-\log p_y$, and its conditional risk is $-\sum_vq_v\log p_v$. Adding and subtracting $-\sum_vq_v\log q_v$ gives
\begin{equation}
\begin{aligned}
-\sum_vq_v\log p_v
&=\underbrace{-\sum_vq_v\log q_v}_{\text{minimum conditional risk}}
+\underbrace{\mathrm{KL}(q\|p)}_{\text{posterior-fitting term}},\\
\mathrm{KL}(q\|p)&=\sum_{v:q_v>0}q_v\log\frac{q_v}{p_v}.
\end{aligned}
\tag{A.4}
\label{eq:app:A.added2}
\end{equation}
We use the convention $0\log0=0$. If $p_v=0<q_v$, both the conditional CE risk and KL divergence are infinite. By nonnegativity of KL divergence, the conditional risk is minimized at $p=q$. Thus CE and the half-Brier loss used here both have the true posterior as their optimal prediction, with KL divergence and half the squared error in probabilities as their respective posterior-fitting terms. Averaging the MSE and CE conditional decompositions over positions and taking expectations with the corresponding training weights yields their population risk decompositions.

\subsection{Error Bounds for Continuous Predictions and Flow Velocities}\label{sec:app:A.2}

Fix $E$ and retain the probability matrices $P,Q$ and readouts $M=PE$, $m^*=QE$. Let $\bar e=V^{-1}\sum_ve_v$ and $E_c=E-\mathbf1_V\bar e^\top$. Since $(P-Q)\mathbf1_V=0$, we have $(P-Q)E=(P-Q)E_c$. Applying the matrix spectral-norm inequality at each position and summing gives
\begin{equation}
\|M-m^*\|_F^2=\|(P-Q)E_c\|_F^2
\le\|E_c\|_{\mathrm{op}}^2\|P-Q\|_F^2.
\tag{A.5}\label{eq:app:A.2a}
\end{equation}
Here $\|E_c\|_{\mathrm{op}}$ is the largest singular value of $E_c$, and $\|\cdot\|_F$ is the Frobenius norm.

Differentiating the Gaussian path in Eq.~\eqref{eq:main:5} and eliminating the Gaussian perturbation gives the conditional path velocity $\tfrac12(\alpha_\gamma^2Z_\gamma-\alpha_\gamma X)$, with the same coefficients as in Eq.~\eqref{eq:main:18}. Conditioning on $Z_\gamma$ and $\gamma$ replaces $X$ by its conditional mean $m^*$ and gives the ideal marginal velocity field \citep{lipman2022flowmatching}:
\begin{equation}
v^*(Z_\gamma,\gamma)
=\mathbb E\!\left[\frac{\mathrm dZ_\gamma}{\mathrm d\gamma}\,\middle|\,Z_\gamma,\gamma\right]
=\frac12(\alpha_\gamma^2Z_\gamma-\alpha_\gamma m^*).
\tag{A.6}\label{eq:app:A.19}
\end{equation}

Subtracting this field from the model velocity and applying Eq.~\eqref{eq:app:A.2a} gives
\begin{equation}
\|v_\theta-v^*\|_F^2
=\frac{\alpha_\gamma^2}{4}\|M-m^*\|_F^2
\le\frac{\alpha_\gamma^2}{4}\|E_c\|_{\mathrm{op}}^2\|P-Q\|_F^2.
\tag{A.7}
\label{eq:app:A.added3}
\end{equation}
Thus, with fixed $E$, the same posterior probability error bounds both continuous readout and velocity errors.

\subsection{Nonidentifiability of Token Probabilities from Continuous Readouts}\label{sec:app:A.4}

We give the linear-algebra argument underlying Section~2.4. Fix $E$ and consider the linear readout \(T_E(p)=E^\top p\). Differences between probability vectors lie in the tangent space \(\mathcal T=\{a\in\mathbb R^V:\mathbf1_V^\top a=0\}\). If $V>d+1$, then

\begin{equation}
\operatorname{rank}\!
\begin{bmatrix}E^\top\\ \mathbf1_V^\top\end{bmatrix}
\le d+1<V,
\qquad
\dim\ker\!
\begin{bmatrix}E^\top\\ \mathbf1_V^\top\end{bmatrix}
\ge V-d-1>0.
\tag{A.8}\label{eq:app:A.13}
\end{equation}

Choose a nonzero null-space vector $a$ and $p_0=\mathbf1_V/V$. For $0<t<1/(V\|a\|_\infty)$, where $\|a\|_\infty=\max_v|a_v|$, both $p_\pm=p_0\pm ta$ have positive entries summing to one, and

\begin{equation}
p_+\ne p_-,
\qquad E^\top p_+=E^\top p_-.
\tag{A.9}\label{eq:app:A.14}
\end{equation}

Thus, distinct identity distributions can yield the same continuous readout even when embeddings are distinct. The two probability vectors give the same continuous reconstruction loss for every true token, since their readouts are identical. Updating $E$ also changes the induced metric. An embedding matrix with identical rows maps every probability residual in this tangent space to zero, as in Eq.~\eqref{eq:main:10}.

Brier conditional excess risk is \(\|p-q\|^2/2\), so it still evaluates identity probability residuals in this null space.

\subsection{Contextual Posteriors and Gaussian Observation Factorization}\label{sec:app:A.5}

Fix $E$ and the noise level $\gamma$. Let $Z_{-\ell}$ contain all noisy observations except position $\ell$, and let $z_\ell$ be the observation column vector at position $\ell$. Given $Y_\ell=v$ and $\gamma$, the local observation is $\alpha_\gamma e_v$ plus independent Gaussian noise and is therefore conditionally independent of $Z_{-\ell}$. Its likelihood is proportional to the kernel $K_{\ell,v}$ in Eq.~\eqref{eq:main:13}, with a candidate-independent density normalization factor. Conditional Bayes\textquotesingle{} rule therefore weights each contextual probability $\Pr(Y_\ell=v\mid Z_{-\ell},\gamma)$ by $K_{\ell,v}$ and normalizes over the vocabulary \citep[Proposition~1]{li2026convergeflow}:

\begin{equation}
q_{\ell,v}(Z,\gamma)=
\frac{\Pr(Y_\ell=v\mid Z_{-\ell},\gamma)\,
K_{\ell,v}}
{\sum_w\Pr(Y_\ell=w\mid Z_{-\ell},\gamma)\,
K_{\ell,w}}.
\tag{A.10}\label{eq:app:A.16}
\end{equation}

Token dependencies are retained in the contextual factor. The true posterior uses the context $Z_{-\ell}$, whereas the base network reads the full $Z$ and is trained to fit the normalized token probabilities.

\subsection{Derivation of the Information-Uniform Noise Schedule}\label{sec:app:A.6}

The trainable schedule estimates the rate of identity-information change using existing denoising predictions and adjusts noise-level sampling accordingly \citep{chen2026langflow,raya2026infonoise}. We derive the information-change rate, sampling density and quantiles, and online measurement. Gumbel fitting, independent parameter updates, and practical sampling are detailed in Appendix~\ref{sec:app:D.3}. Derivatives with respect to the noise coordinate are computed for a Gaussian channel with fixed $E$.

\textbf{Information-change rate.} Let \(\mathsf H(\gamma)=H(Y\mid Z_\gamma)/L\) be joint conditional entropy per token, measured in nats. Here $H$ is Shannon entropy and $Y$ is the complete sequence. Invertibly rescale the Gaussian observation as

\begin{equation}
W_\lambda:=\frac{Z_\gamma}{\sigma_\gamma}
=\sqrt\lambda X+\varepsilon,
\qquad \lambda=\frac{\alpha_\gamma^2}{\sigma_\gamma^2}=e^{-\gamma}.
\tag{A.11}
\label{eq:app:A.added4}
\end{equation}

For fixed $\gamma$, $Z_\gamma$ and $W_\lambda$ are invertible rescalings and contain the same label information. Since $X$ is a deterministic function of $Y$ and \(Y\to X\to W_\lambda\) is a Markov chain, \(I(Y;W_\lambda)=I(X;W_\lambda)\). Here $I$ denotes Shannon mutual information in nats, so \(\mathsf H=(H(Y)-I(X;W_\lambda))/L\). A fixed finite embedding set has finite second moments. Applying the Gaussian-channel I--MMSE identity to the sequence matrices flattened into $Ld$-dimensional vectors gives \citep{guo2005immse}:

\begin{equation}
\frac{\partial}{\partial\lambda}I(X;W_\lambda)
=\frac12\mathbb E\left[
\|X-\mathbb E[X\mid W_\lambda]\|_F^2\right].
\tag{A.12}
\label{eq:app:A.added5}
\end{equation}

At the corresponding $\gamma$, the inner conditional mean is $m^*$ from Appendix~\ref{sec:app:A.2}. The chain rule with \(\mathrm d\lambda/\mathrm d\gamma=-e^{-\gamma}\) gives the per-token conditional-entropy change rate

\begin{equation}
g(\gamma):=\frac{\mathrm d\mathsf H(\gamma)}{\mathrm d\gamma}
=\frac{e^{-\gamma}}{2L}
\mathbb E[\|X-m^*(Z_\gamma,\gamma)\|_F^2\mid\gamma]\ge0.
\tag{A.13}\label{eq:app:A.9}
\end{equation}

The function $g(\gamma)$ measures information change per unit change in the noise coordinate; the factor $e^{-\gamma}$ follows from the SNR reparameterization.

\textbf{From information change to a noise distribution.} Fix a finite window $[\gamma_{\mathrm{lo}},\gamma_{\mathrm{hi}}]$ and let $\Delta\mathsf H=\mathsf H(\gamma_{\mathrm{hi}})-\mathsf H(\gamma_{\mathrm{lo}})=\int_{\gamma_{\mathrm{lo}}}^{\gamma_{\mathrm{hi}}}g(s)\,\mathrm ds$. If $X$ is not almost surely constant, its conditional variance under finite Gaussian noise is positive, so $g(\gamma)>0$ and $\Delta\mathsf H>0$. Normalizing gives the sampling density and CDF:

\begin{equation}
\pi^*(\gamma)=\frac{g(\gamma)}{\Delta\mathsf H},
\qquad F^*(\gamma)=\int_{\gamma_{\mathrm{lo}}}^{\gamma}\pi^*(s)\,\mathrm ds
=\frac{\mathsf H(\gamma)-\mathsf H(\gamma_{\mathrm{lo}})}{\Delta\mathsf H}.
\tag{A.14}\label{eq:app:A.10}
\end{equation}

Writing $u=F^*(\gamma)\in[0,1]$, we obtain

\begin{equation}
\mathsf H\big((F^*)^{-1}(u)\big)
=\mathsf H(\gamma_{\mathrm{lo}})+u\,\Delta\mathsf H.
\tag{A.15}
\label{eq:app:A.added6}
\end{equation}

Uniformly spaced quantiles therefore correspond to equal conditional-entropy increments. Sampling from this ideal information-uniform density and using decreasing quantiles for generation concentrates evaluations where information changes rapidly. The implementation approximates this density by fitting a Gumbel distribution to the online measurements below.

\textbf{Online measurement.} Approximate the unknown conditional mean $m^*$ with model readout $M$ and define

\begin{equation}
\widehat e=\frac{\|X-M\|_F^2}{Ld},\qquad
\widehat g=\frac{e^{-\gamma}}{2L}\|X-M\|_F^2.
\tag{A.16}\label{eq:app:A.11}
\end{equation}

This measurement supplies the schedule target; the update rule is given in Appendix~\ref{sec:app:D.3}.

To relate the measurement to the ideal information density, write \(X-M=(X-m^*)+(m^*-M)\). The conditional-mean identity \(\mathbb E[X-m^*\mid Z_\gamma,\gamma,\Omega]=0\) makes the cross term vanish in expectation. Under the independence of prediction randomness from data and observation noise specified in Appendix~\ref{sec:app:A.1},

\begin{equation}
\begin{aligned}
\mathbb E[\widehat e\mid\gamma]
&=\frac{\mathbb E[\|X-m^*\|_F^2\mid\gamma]}{Ld}
+\frac{\mathbb E[\|M-m^*\|_F^2\mid\gamma]}{Ld},\\
\mathbb E[\widehat g\mid\gamma]
&=g(\gamma)
+\frac{e^{-\gamma}}{2L}
\mathbb E[\|M-m^*\|_F^2\mid\gamma].
\end{aligned}
\tag{A.17}\label{eq:app:A.12}
\end{equation}

We can further connect measurement bias to identity supervision. Following Eq.~\eqref{eq:main:23}, let \(\varepsilon_B(\gamma):=\mathbb E[\|P-Q\|_F^2/(2L)\mid\gamma]\) denote unweighted Brier excess risk averaged over positions at fixed noise. The expectation covers data, Gaussian perturbations, and independent prediction randomness at that level. With fixed $E$, Eq.~\eqref{eq:app:A.12} and the centered-embedding spectral bound in Eq.~\eqref{eq:app:A.2a} give

\begin{equation}
0\le\mathbb E[\widehat g\mid\gamma]-g(\gamma)
=\frac{e^{-\gamma}}{2L}\mathbb E[\|M-m^*\|_F^2\mid\gamma]
\le e^{-\gamma}\|E_c\|_{\mathrm{op}}^2\varepsilon_B(\gamma).
\tag{A.18}\label{eq:app:A.12a}
\end{equation}

This bound controls the nonnegative bias of the information-change measurement by posterior squared error; when $M=m^*$, the measurement equals the true information-change rate in expectation.

\section{Risk Analysis of Frozen-Embedding and Joint Continuation}\label{sec:app:B}

\subsection{Risk Comparison between Frozen-Embedding and Joint Continuation}\label{sec:app:B.1}
Throughout this appendix, fix the data distribution, noise distribution $\pi$, and positive weight $w$, under the finite-risk setting of Appendix~\ref{sec:app:A.1}. Let $E_0$ be pretrained embeddings and let \((\theta_{\mathrm f},E_0)\) and \((\theta_{\mathrm j},E_{\mathrm j})\) denote the frozen and joint continuation results, evaluated using the same population risk as in the main text. Each embedding matrix defines its own noisy observations and true posteriors.

Define changes in minimum risk and posterior fitting error as
\begin{equation}
\Delta\mathcal R^*=\mathcal R^*(E_{\mathrm j})-\mathcal R^*(E_0),\qquad
\Delta\mathcal E=\mathcal E(\theta_{\mathrm j},E_{\mathrm j})-\mathcal E(\theta_{\mathrm f},E_0).
\tag{B.1}\label{eq:app:B.1}
\end{equation}
Since \(\mathcal R=\mathcal R^*+\mathcal E\),
\begin{equation}
\mathcal R(\theta_{\mathrm j},E_{\mathrm j})-\mathcal R(\theta_{\mathrm f},E_0)
=\Delta\mathcal R^*+\Delta\mathcal E.
\tag{B.2}\label{eq:app:B.2}
\end{equation}
Lower risk for the joint result can reflect improvement in both components or improvement in one that outweighs deterioration in the other.

\subsection{Remaining Embedding Improvement and Bounds on Risk Gains}\label{sec:app:B.2}
Let $\mathscr C$ be the set of $V\times d$ embedding matrices produced by the RowNorm parameterization in Eq.~\eqref{eq:main:11}, containing both $E_0$ and $E_{\mathrm j}$. Define the infimum of the minimum risk and the remaining improvement available from the pretrained embeddings:
\begin{equation}
\mathcal R^*_{\min}:=\inf_{E\in\mathscr C}\mathcal R^*(E),\qquad
\delta_{\mathrm{rep}}:=\mathcal R^*(E_0)-\mathcal R^*_{\min}.
\tag{B.3}\label{eq:app:B.3}
\end{equation}
Since $E_0\in\mathscr C$, $\delta_{\mathrm{rep}}\ge0$. It is the largest further reduction in minimum identity risk available by adjusting $E$. Motivated by the observation that associations between token embeddings and semantic features emerge early and stabilize during pretraining \citep{papadimitriou2025vocabulary}, we consider sufficiently pretrained embeddings $E_0$ and assume that $\delta_{\mathrm{rep}}$ is small.

For the particular joint result, $E_{\mathrm j}\in\mathscr C$ implies $\mathcal R^*(E_{\mathrm j})\ge\mathcal R^*_{\min}$. Combining this with the definitions in Eqs.~\eqref{eq:app:B.1} and~\eqref{eq:app:B.3}, we obtain

\begin{equation}
\begin{aligned}
\Delta\mathcal R^*
&=\mathcal R^*(E_{\mathrm j})-\mathcal R^*(E_0)\\
&\ge\mathcal R^*_{\min}-\mathcal R^*(E_0)
=-\delta_{\mathrm{rep}}.
\end{aligned}
\tag{B.4}
\label{eq:app:B.added8}
\end{equation}

Substituting this lower bound into the exact risk difference in Eq.~\eqref{eq:app:B.2} gives
\begin{equation}
\mathcal R(\theta_{\mathrm j},E_{\mathrm j})-\mathcal R(\theta_{\mathrm f},E_0)
\ge\Delta\mathcal E-\delta_{\mathrm{rep}}.
\tag{B.5}\label{eq:app:B.5}
\end{equation}
If $\Delta\mathcal E>\delta_{\mathrm{rep}}$, the right-hand side is strictly positive. Thus, when the existing embeddings have little remaining room for improvement and the joint result has a higher posterior fitting error than the frozen result by more than this amount, the frozen result has lower total identity risk.

\subsection{Competing Risk Components under Joint Updates}\label{sec:app:B.3}

Under the same fixed distributions and weight, assume normalization and both risk components are differentiable in the region considered, and analyze continuous-time gradient flow. Let $\zeta=\operatorname{vec}(A)\in\mathbb R^{Vd}$, where $\operatorname{vec}$ flattens a matrix in a fixed order. Parameter gradients use the Euclidean inner product $\langle\cdot,\cdot\rangle$ and norm $\|\cdot\|$. Minimum risk depends on $\zeta$ through $E$ and is independent of $\theta$, whereas fitting error depends on both. Derivatives with respect to $\zeta$ include changes in noisy observations and true posteriors induced by $E$. Write
\begin{equation}
g_{\mathrm{rep}}=\nabla_\zeta\mathcal R^*,\qquad
g_{\mathrm{fit}}=\nabla_\zeta\mathcal E,\qquad
g_\theta=\nabla_\theta\mathcal E.
\tag{B.6}
\label{eq:app:B.added9}
\end{equation}
Let $\dot\zeta=-\eta_A(g_{\mathrm{rep}}+g_{\mathrm{fit}})$ and $\dot\theta=-\eta_\theta g_\theta$, where $\eta_A,\eta_\theta>0$ are update-rate factors and dots denote derivatives with respect to optimization time. The chain rule gives $\dot{\mathcal R}^*=\langle g_{\mathrm{rep}},\dot\zeta\rangle$ and $\dot{\mathcal E}=\langle g_{\mathrm{fit}},\dot\zeta\rangle+\langle g_\theta,\dot\theta\rangle$. Substituting the update directions and using $\dot{\mathcal R}=\dot{\mathcal R}^*+\dot{\mathcal E}$ yields
\begin{equation}
\begin{aligned}
\dot{\mathcal R}&=-\eta_\theta\|g_\theta\|^2-\eta_A\|g_{\mathrm{rep}}+g_{\mathrm{fit}}\|^2\le0,\\
\dot{\mathcal R}^*&=-\eta_A\bigl(\|g_{\mathrm{rep}}\|^2+\langle g_{\mathrm{rep}},g_{\mathrm{fit}}\rangle\bigr),\\
\dot{\mathcal E}&=-\eta_\theta\|g_\theta\|^2-\eta_A\bigl(\|g_{\mathrm{fit}}\|^2+\langle g_{\mathrm{rep}},g_{\mathrm{fit}}\rangle\bigr).
\end{aligned}
\tag{B.7}\label{eq:app:B.conflict}
\end{equation}
By Eq.~\eqref{eq:app:B.conflict}, if $\langle g_{\mathrm{rep}},g_{\mathrm{fit}}\rangle\ge0$, neither component increases. If $\langle g_{\mathrm{rep}},g_{\mathrm{fit}}\rangle<-\|g_{\mathrm{rep}}\|^2$, positivity of $\eta_A$ gives $\dot{\mathcal R}^*>0$. Since $\dot{\mathcal R}\le0$, differentiating the risk decomposition directly yields
\begin{equation}
\dot{\mathcal E}
=\dot{\mathcal R}-\dot{\mathcal R}^*
\le-\dot{\mathcal R}^*<0.
\tag{B.8}
\label{eq:app:B.added10}
\end{equation}
Under this condition, the decrease in fitting error is at least as large as the increase in minimum risk, so total risk remains nonincreasing. With frozen embeddings, $\dot\zeta=0$, so $\dot{\mathcal R}^*=0$ and $\dot{\mathcal E}=-\eta_\theta\|g_\theta\|^2$: risk reduction is concentrated on posterior fitting within the fixed channel.

\section{Endpoint Convergence of Continuous Trajectories}\label{sec:app:C}

Starting from the conditional-risk optimum of TIBF, this appendix establishes the endpoint distribution under optimal posteriors, local convergence of the actual sampling updates, and decoding agreement at finite noise. During generation, $E$ remains fixed at either the pretrained embeddings used in frozen continuation or the final embeddings learned by joint continuation.

\subsection{Setting and Notation}\label{sec:app:C.1}

Retain the token sequence $Y$ and embedding sequence $X$ from the main text, and write

\begin{equation}
D=Ld,
\qquad \mu_X=\operatorname{Law}(X).
\tag{C.1}\label{eq:app:C.1}
\end{equation}

Here \(Y\sim P_{\rm data}\), $D$ is the number of coordinates in the entire continuous sequence, and $\mu_X$ is the distribution of $X$. We identify matrices with their $D$-dimensional flattened representations and use the corresponding Frobenius norm. For fixed vocabulary and length,

\begin{equation}
\mathcal S=\{X(y):P_{\rm data}(y)>0\}
\subseteq\{(e_{y_1},\ldots,e_{y_L})^\top:y\in[V]^L\}
\tag{C.2}\label{eq:app:C.2}
\end{equation}

Here $X(y)$ is the embedding matrix associated with token sequence $y$, and $\mathcal S$ is the finite set of possible embedding sequences.

Use the Gaussian path in Eq.~\eqref{eq:main:5}, with noise independent of $Y$. To analyze the endpoint, define the noise-to-signal ratio and signal-scaled state

\begin{equation}
h=\frac{\sigma_\gamma}{\alpha_\gamma}=e^{\gamma/2},
\qquad U_h=\frac{Z_\gamma}{\alpha_\gamma}.
\tag{C.3}\label{eq:app:C.4}
\end{equation}

Denote the target marginal of the corruption path at $h$ by

\begin{equation}
\nu_h=\operatorname{Law}(X+h\varepsilon).
\tag{C.4}\label{eq:app:C.5}
\end{equation}

The clean endpoint is $h\downarrow0$, equivalently $\gamma\to-\infty$. Below we show that the ideal generation field transports the marginals $\nu_h$.

For fixed $h$, rescaling is invertible. Let $u$ denote a value of the rescaled state and define

\begin{equation}
\begin{aligned}
q^h_{\ell,v}(u)
&=\Pr(Y_\ell=v\mid X+h\varepsilon=u),\\
m_h^*(u)&=\mathbb E[X\mid X+h\varepsilon=u]
=Q^h(u)E.
\end{aligned}
\tag{C.5}\label{eq:app:C.6}
\end{equation}

Here $Q^h(u)=(q^h_{\ell,v}(u))_{\ell,v}$ is the positionwise posterior matrix, $q^h_\ell(u)$ is the transpose of its $\ell$th row, and $m_h^*(u)$ is the continuous conditional mean.

\subsection{Brier Optimality and the Ideal Generation Field}\label{sec:app:C.2}

By Eq.~\eqref{eq:main:23}, the unique Brier-optimal prediction is the true posterior $q^h_\ell(u)$, giving the continuous prediction $m_h^*=Q^hE$. Substituting it into Eq.~\eqref{eq:main:18} and using $\alpha_\gamma'/\alpha_\gamma=-\sigma_\gamma^2/2$ gives

\begin{equation}
\frac{\mathrm dU_h}{\mathrm d\gamma}
=\frac{v^*}{\alpha_\gamma}
-\frac{\alpha_\gamma'}{\alpha_\gamma}U_h
=\frac12\bigl((\alpha_\gamma^2+\sigma_\gamma^2)U_h-m_h^*(U_h)\bigr)
=\frac12(U_h-m_h^*(U_h)),
\tag{C.6}
\label{eq:app:C.added12}
\end{equation}
where $v^*$ is the velocity obtained by replacing $M$ with $m_h^*$. Since $\mathrm dh/\mathrm d\gamma=h/2$, we obtain

\begin{equation}
\frac{\mathrm dU_h}{\mathrm dh}
=b^*(U_h,h),
\qquad b^*(u,h)=\frac{u-m_h^*(u)}h.
\tag{C.7}\label{eq:app:C.8}
\end{equation}

Generation integrates in decreasing $h$; using $-\log h$ as time gives direction $m_h^*-U_h$. Write $\nu_h(u)$ for the density of the target distribution. Equation~\eqref{eq:app:C.5} gives
\begin{equation}
\nu_h(u)=\sum_{x\in\mathcal S}\mu_X(\{x\})
\frac{\exp[-\|u-x\|^2/(2h^2)]}{(2\pi h^2)^{D/2}},
\tag{C.8}
\label{eq:app:C.added13}
\end{equation}
where $\mu_X(\{x\})$ is the probability of embedding sequence $x$. Differentiation with respect to $u$ multiplies each component by $(x-u)/h^2$; its normalized weighted average is $(m_h^*(u)-u)/h^2$. Differentiating each Gaussian component with respect to $h$ and summing gives $\partial_h\nu_h=h\Delta_u\nu_h$. Therefore,

\begin{equation}
\nabla_u\log\nu_h(u)=\frac{m_h^*(u)-u}{h^2},
\qquad
\partial_h\nu_h=h\Delta_u\nu_h
=-\nabla_u\!\cdot(\nu_h b^*).
\tag{C.9}\label{eq:app:C.9}
\end{equation}

Here $\nabla_u$, $\nabla_u\!\cdot$, and $\Delta_u$ denote the gradient, divergence, and Laplacian with respect to $u$. The last equality uses $\nu_hb^*=-h\nabla_u\nu_h$. This continuity equation establishes compatibility between the velocity induced by Brier-optimal predictions and the Gaussian marginal path \citep{lipman2022flowmatching}.

\subsection{Endpoint Convergence under Ideal Posteriors}\label{sec:app:C.3}

\textbf{Theorem 1 (Endpoint convergence under ideal posteriors).} Fix $E$, a finite sequence length $L$, and a finite vocabulary size $V$. For finite $h_0>0$, assume TIBF's predicted probabilities equal the true posterior in Eq.~\eqref{eq:app:C.6} for every $0<h\le h_0$, and solve Eq.~\eqref{eq:app:C.8} exactly from $U_{h_0}\sim\nu_{h_0}$. Then the trajectory converges almost surely to a random limit $U_*\in\mathcal S$, with $\operatorname{Law}(U_*)=\mu_X$, and

\begin{equation}
\left(\mathbb E\|U_h-U_*\|_F^2\right)^{1/2}
\le\sqrt{Ld}\,h.
\tag{C.10}\label{eq:app:C.10}
\end{equation}

If embeddings are pairwise distinct, $U_*$ uniquely identifies a token sequence $Y^*\sim P_{\rm data}$, and
\begin{equation}
Z_\gamma\longrightarrow X(Y^*),
\qquad p_\ell\longrightarrow\delta_{Y^*_\ell},
\qquad m_\ell\longrightarrow e_{Y^*_\ell}.
\tag{C.11}\label{eq:app:C.11}
\end{equation}

\subsubsection{Unique Flow and Marginal Distributions}\label{sec:app:C.3.1}

A finite Gaussian mixture has a smooth, everywhere-positive density for $h>0$. Flatten the state and $X$ into $D$-dimensional vectors. Applying the quotient rule to the Gaussian-weighted conditional mean gives
\begin{equation}
\nabla_u m_h^*(u)=h^{-2}\operatorname{Cov}(X\mid X+h\varepsilon=u).
\tag{C.12}
\label{eq:app:C.added14}
\end{equation}
Here $\operatorname{Cov}$ denotes conditional covariance. Finite support bounds this covariance uniformly. Hence, for every $0<a<h_0$, both $m_h^*$ and $b^*=(u-m_h^*)/h$ are globally Lipschitz in $u$ on $[a,h_0]$, and the velocity has at most linear growth.

Equation~\eqref{eq:app:C.8} therefore defines a unique flow on each such interval, and its pushforward distributions satisfy the continuity equation. Under the stated Lipschitz and growth conditions, this equation also has a unique solution. Equation~\eqref{eq:app:C.9} and the common initial distribution thus imply that the flow marginal at $h$ is $\nu_h$. Uniqueness makes the solutions for all $a>0$ compatible, defining the same trajectory on $(0,h_0]$.

\subsubsection{Trajectory Limits and Mean-Square Error Bounds}\label{sec:app:C.3.2}

Taking conditional expectations in $X+h\varepsilon-X=h\varepsilon$ given the observation yields

\begin{equation}
\frac{u-m_h^*(u)}h
=\mathbb E[\varepsilon\mid X+h\varepsilon=u].
\tag{C.13}\label{eq:app:C.12}
\end{equation}

Under these marginals, the noise residual is orthogonal in expectation to its conditional mean, so
\begin{equation}
\begin{aligned}
\mathbb E\|b^*(U_h,h)\|_F^2
&=D-\mathbb E\|\varepsilon-\mathbb E[\varepsilon\mid X+h\varepsilon]\|_F^2
\le D,\\
\mathbb E\|b^*(U_h,h)\|_F&\le\sqrt D.
\end{aligned}
\tag{C.14}\label{eq:app:C.13}
\end{equation}

Since path length is nonnegative, exchanging integration and expectation gives
\begin{equation}
\begin{aligned}
\mathbb E\!\left[\int_0^{h_0}
\left\|\frac{\mathrm dU_h}{\mathrm dh}\right\|_F\mathrm dh\right]
&=\int_0^{h_0}\mathbb E\|b^*(U_h,h)\|_F\,\mathrm dh\\
&\le\sqrt D\,h_0<\infty.
\end{aligned}
\tag{C.15}\label{eq:app:C.14}
\end{equation}

Finite expected path length implies that the path length is finite almost surely. The remaining length on $(0,h]$ vanishes as $h\downarrow0$, so the distance between any two points in this tail also vanishes. Hence the trajectory admits a finite limit $U_*$.

For $0<a<h$, write the trajectory difference as an integral of velocity. Cauchy--Schwarz and Eq.~\eqref{eq:app:C.13} give
\begin{equation}
\begin{aligned}
\mathbb E\|U_h-U_a\|_F^2
&=\mathbb E\left\|\int_a^h b^*(U_s,s)\,\mathrm ds\right\|_F^2\\
&\le(h-a)\int_a^h\mathbb E\|b^*(U_s,s)\|_F^2\,\mathrm ds
\le D(h-a)^2.
\end{aligned}
\tag{C.16}\label{eq:app:C.15}
\end{equation}

Thus convergence also holds in mean square, with the same almost-sure limit. Taking $a\downarrow0$ gives Eq.~\eqref{eq:app:C.10}. After normalization by the square root of the number of coordinates, the root-mean-square error is bounded by $h$.

\subsubsection{Endpoint Distribution}\label{sec:app:C.3.3}

By Eq.~\eqref{eq:app:C.5}, the target distribution $\nu_h$ converges weakly to $\mu_X$ as $h$ vanishes. The preceding proof establishes that the generated state has distribution $\nu_h$ and converges almost surely to $U_*$, so the same distribution also converges weakly to $\operatorname{Law}(U_*)$. Uniqueness of the limit gives

\begin{equation}
\operatorname{Law}(U_*)=\mu_X,
\qquad \Pr(U_*\in\mathcal S)=1.
\tag{C.17}\label{eq:app:C.16}
\end{equation}

Since $\alpha_\gamma=(1+h^2)^{-1/2}\to1$, we also have $Z_\gamma=\alpha_\gamma U_h\to U_*$.

\subsubsection{Token Identity Recovery}\label{sec:app:C.3.4}

With pairwise distinct embeddings, $y\mapsto X(y)$ is injective. Suppose the trajectory limit is $x^*=X(y^*)$ and consider the full sequence posterior

\begin{equation}
q_h^{\mathrm{seq}}(y\mid u)=
\frac{P_{\rm data}(y)\exp[-\|u-X(y)\|_F^2/(2h^2)]}
{\sum_{y'}P_{\rm data}(y')\exp[-\|u-X(y')\|_F^2/(2h^2)]}.
\tag{C.18}\label{eq:app:C.18}
\end{equation}

Its positionwise marginals are Eq.~\eqref{eq:app:C.6}; the full sequence posterior retains dependencies across positions. If $x^*$ is the only support point, the claim is immediate. Otherwise, let $\Delta_*>0$ be its minimum distance to other support points. Once $\|U_h-x^*\|_F\le\Delta_*/4$, write $a_y=\|X(y)-x^*\|_F\ge\Delta_*$. Expanding the squared distances and applying Cauchy--Schwarz gives

\begin{equation}
\|U_h-X(y)\|_F^2-\|U_h-x^*\|_F^2
\ge a_y^2-2a_y\|U_h-x^*\|_F\ge\frac{\Delta_*^2}{2}.
\tag{C.19}\label{eq:app:C.19}
\end{equation}

Combining Eqs.~\eqref{eq:app:C.18} and \eqref{eq:app:C.19} bounds the total posterior mass of sequences other than $y^*$ by \(\frac{1-P_{\rm data}(y^*)}{P_{\rm data}(y^*)}\allowbreak\times\allowbreak
\exp[-\Delta_*^2/(4h^2)]\), which vanishes as $h\downarrow0$. The full posterior therefore concentrates on $y^*$; each positionwise posterior and ideal prediction converges to its one-hot label, and the readout converges to the same embedding, proving Eq.~\eqref{eq:app:C.11}. This completes the proof.

\subsection{Local Convergence under Gaussian-Weighted Predictions}\label{sec:app:C.5}

\subsubsection{Convex-Combination Form of Sampling Updates}\label{sec:app:C.5.1}

Let $i$ index sampling nodes, with strictly decreasing $h_i>0$, $\gamma_i=2\log h_i$, and $\alpha_i=(1+h_i^2)^{-1/2}$. Write $U_i$ for the rescaled state, $Z_i=\alpha_iU_i$ for the original state, $P_i$ for the token-probability matrix, and $M_i=P_iE$ for the continuous prediction at this node. Let $z_{i,\ell}$ denote the transpose of row $\ell$ of $Z_i$. For the actual predictor, replace $m_h^*$ above by $M=PE$. Holding the current prediction $M_i$ fixed within $[h_{i+1},h_i]$ gives
\begin{equation}
\frac{\mathrm dU}{\mathrm dh}=\frac{U-M_i}{h}
\quad\Longrightarrow\quad
\frac{\mathrm d}{\mathrm dh}\left(\frac{U-M_i}{h}\right)=0.
\tag{C.20}
\label{eq:app:C.added15}
\end{equation}
Evaluating at the two interval endpoints yields the update used by the data-prediction sampler:

\begin{equation}
U_{i+1}=\chi_iU_i+(1-\chi_i)M_i,
\qquad
\chi_i=\frac{h_{i+1}}{h_i}\in(0,1).
\tag{C.21}\label{eq:app:C.23}
\end{equation}

Each row of $M_i=P_iE$, written as a column vector $m_{i,\ell}$, lies in the convex hull of the embeddings. Likewise, $U_{i,\ell}$ denotes the transpose of row $\ell$ of $U_i$. Thus $U_{i,\ell}$ remains in the convex hull of its initial state and the embedding set.

Throughout this local analysis, token embeddings are nonzero and have equal norm, the Gaussian branch is fully enabled, and no additional guidance extrapolation is applied. Let $f_i=f_\theta(Z_i,\gamma_i,S_i)$ be the base-score matrix and $S_i$ the current self-conditioning cache. Equation~\eqref{eq:main:15} becomes

\begin{equation}
p_{i,\ell v}
=\frac{\omega_{i,\ell v}
\exp[-\|U_{i,\ell}-e_v\|^2/(2h_i^2)]}
{\sum_w\omega_{i,\ell w}
\exp[-\|U_{i,\ell}-e_w\|^2/(2h_i^2)]},
\qquad \omega_i=\operatorname{softmax}(f_i).
\tag{C.22}\label{eq:app:C.24}
\end{equation}

Indeed, $\alpha_\gamma^2/\sigma_\gamma^2=1/h^2$, and equal-norm embeddings satisfy
\begin{equation}
-\frac{\|U_{i,\ell}-e_v\|^2}{2h_i^2}
=\frac{U_{i,\ell}^{\top}e_v}{h_i^2}
-\frac{\|U_{i,\ell}\|^2+\|e_1\|^2}{2h_i^2}.
\tag{C.23}
\label{eq:app:C.added16}
\end{equation}
The second term is shared across candidates and cancels under probability normalization, so Eq.~\eqref{eq:app:C.24} is equivalent to the Gaussian observation scores in the main text. Softmax operates over the vocabulary dimension; $p_{i,\ell v}$ and $\omega_{i,\ell v}$ are the predicted probability and base weight for candidate $v$ at node $i$ and position $\ell$.

\subsubsection{Conditions and Statement}\label{sec:app:C.5.2}

Assume $V\ge2$ and pairwise distinct embeddings. Define the minimum separation $\Delta$, neighborhood radius $R$, and diameter $D_E$ by

\begin{equation}
\Delta=\min_{v\ne w}\|e_v-e_w\|>0,
\quad R=\Delta/4,
\quad D_E=\max_{v,w}\|e_v-e_w\|.
\tag{C.24}\label{eq:app:C.25}
\end{equation}

\textbf{Theorem 2 (local endpoint convergence).} Under the setting above, consider Eq.~\eqref{eq:app:C.23} on an infinite decreasing grid $h_i\downarrow0$. Fix a position $\ell$ and candidate token $j$. Suppose that, from some index $i_0$ onward:

\begin{itemize}
\tightlist
\item
  The state at the start of this tail satisfies $\|U_{i_0,\ell}-e_j\|\le R$;
\item
  Whenever tail states lie in this ball, the actual network and cache satisfy $f_{i,\ell v}-f_{i,\ell j}\le B_f$ for every candidate $v$, with a finite constant $B_f\ge0$ independent of $i$;
\item
  Define the error envelope $a(h)$ below and require the starting noise level to satisfy
\end{itemize}

\begin{equation}
\begin{aligned}
 a(h)&:=D_E(V-1)\exp\!\left(B_f-\frac{\Delta^2}{4h^2}\right),\quad h>0,\\
 a(h_{i_0})&\le R.
\end{aligned}
  \tag{C.25}\label{eq:app:C.26}
\end{equation}

All subsequent states satisfy $\|U_{i,\ell}-e_j\|\le R$, and

\begin{equation}
U_{i,\ell}\to e_j,
\quad z_{i,\ell}\to e_j,
\quad m_{i,\ell}\to e_j,
\quad p_{i,\ell j}\to1.
\tag{C.26}\label{eq:app:C.27}
\end{equation}

\subsubsection{Proof of Local Attraction}\label{sec:app:C.5.3}

If $\|u-e_j\|\le R$, take any $v\ne j$ and write $c=\|e_v-e_j\|\ge\Delta$. Expanding squared distances yields

\begin{equation}
\|u-e_v\|^2-\|u-e_j\|^2
\ge c^2-2Rc\ge c^2/2\ge\Delta^2/2.
\tag{C.27}\label{eq:app:C.28}
\end{equation}

Equation~\eqref{eq:app:C.24} gives \(p_{i,\ell v}/p_{i,\ell j}=\exp(f_{i,\ell v}-f_{i,\ell j})\exp[-(\|U_{i,\ell}-e_v\|^2-\|U_{i,\ell}-e_j\|^2)/(2h_i^2)]\). Combining the distance and score bounds with \(m_{i,\ell}-e_j=\sum_{v\ne j}p_{i,\ell v}(e_v-e_j)\) yields

\begin{equation}
1-p_{i,\ell j}
\le\sum_{v\ne j}\frac{p_{i,\ell v}}{p_{i,\ell j}}
\le(V-1)\exp\!\left(B_f-\frac{\Delta^2}{4h_i^2}\right),
\qquad
\|m_{i,\ell}-e_j\|\le a(h_i).
\tag{C.28}\label{eq:app:C.29}
\end{equation}

Because $h$ decreases, $a(h_i)\le a(h_{i_0})\le R$. Whenever $U_{i,\ell}$ lies in the ball, Eq.~\eqref{eq:app:C.29} gives $\|m_{i,\ell}-e_j\|\le R$. The convex update therefore satisfies
\begin{equation}
\|U_{i+1,\ell}-e_j\|
\le\chi_i\|U_{i,\ell}-e_j\|+(1-\chi_i)\|m_{i,\ell}-e_j\|
\le R.
\tag{C.29}
\label{eq:app:C.added17}
\end{equation}
Induction establishes neighborhood invariance and preserves these estimates at every subsequent node.

Define the state residual $H_i=U_{i,\ell}-e_j$ and prediction residual $\xi_i=m_{i,\ell}-e_j$. Then

\begin{equation}
H_{i+1}=\chi_iH_i+(1-\chi_i)\xi_i,
\qquad \|\xi_i\|\le a(h_i)\to0.
\tag{C.30}\label{eq:app:C.30}
\end{equation}

For $N>K\ge i_0$, expand the recursion using $\prod_{i=K}^{N-1}\chi_i=h_N/h_K$:
\begin{equation}
H_N=\frac{h_N}{h_K}H_K+
\sum_{i=K}^{N-1}\frac{h_N}{h_{i+1}}(1-\chi_i)\xi_i.
\tag{C.31}
\label{eq:app:C.added18}
\end{equation}
The summation coefficients are nonnegative and sum to $1-h_N/h_K$, so

\begin{equation}
\|H_N\|
\le\frac{h_N}{h_K}\|H_K\|
+\left(1-\frac{h_N}{h_K}\right)
\sup_{i\ge K}\|\xi_i\|.
\tag{C.32}\label{eq:app:C.31}
\end{equation}

Taking $N\to\infty$ and then $K\to\infty$ makes the right-hand side vanish, so $U_{i,\ell}\to e_j$. Equation~\eqref{eq:app:C.29} gives probability and readout convergence; $\alpha_i\to1$ gives convergence of $Z$. This completes the proof.

\subsubsection{Bounded Base Scores}\label{sec:app:C.5.4}

Noise conditioning uses bounded sine and cosine features followed by an MLP with finite weights; dropout is disabled during generation. Equation~\eqref{eq:app:C.23} bounds rescaled states, so original states $Z_i=\alpha_iU_i$ are bounded as well. The self-conditioning cache is initialized at a finite value and subsequently updated as $S_{i+1}=M_i$, so it is also bounded. The closure of these input features is compact in finite dimensions. With finite parameters and normalization with positive stabilizers, the backbone is continuous on this set, so its base scores and their differences have uniform finite bounds.

Together with vocabulary size and minimum embedding separation, this bound determines the starting-noise threshold in Eq.~\eqref{eq:app:C.26}.

\subsection{Decoding Agreement at Finite Noise}\label{sec:app:C.8}

Retain the equal-norm, distinct embeddings and Gaussian prediction structure above. For one sequence, let $j$ be the unique nearest embedding to the scaled state $U_{i,\ell}=z_{i,\ell}/\alpha_i$ at position $\ell$ and node $i$. Define the distance gap and base-score advantage

\begin{equation}
\begin{aligned}
\Delta_{i,\ell}^{\mathrm{dist}}
&=\min_{v\ne j}\!
\left(\|U_{i,\ell}-e_v\|^2-\|U_{i,\ell}-e_j\|^2\right)>0,
\\
B_{i,\ell}^{\mathrm{logit}}
&=\max_v(f_{i,\ell v}-f_{i,\ell j})\ge0.
\end{aligned}
\tag{C.33}\label{eq:app:C.34}
\end{equation}

Let $s_{i,\ell v}$ denote the total score for candidate $v$. Gaussian observation scoring gives, for any $v\ne j$,

\begin{equation}
s_{i,\ell j}-s_{i,\ell v}
=f_{i,\ell j}-f_{i,\ell v}
+\frac{\|U_{i,\ell}-e_v\|^2-\|U_{i,\ell}-e_j\|^2}{2h_i^2}.
\tag{C.34}\label{eq:app:C.35}
\end{equation}

If $\Delta_{i,\ell}^{\mathrm{dist}}/(2h_i^2)>B_{i,\ell}^{\mathrm{logit}}$, token $j$ has the strictly largest total logit, and the decoding rules agree. Sort all normalized distances $\|U_{i,\ell}-e_v\|/(\sqrt d\,h_i)$ and denote the smallest two by $D_{i,\ell,(1)}$ and $D_{i,\ell,(2)}$. The condition is equivalent to

\begin{equation}
\frac{\Delta_{i,\ell}^{\mathrm{dist}}}{2h_i^2}
=\frac d2
\left[(D_{i,\ell,(2)})^2-(D_{i,\ell,(1)})^2\right]
>B_{i,\ell}^{\mathrm{logit}}.
\tag{C.35}\label{eq:app:C.36}
\end{equation}

Thus, probability decoding and state nearest-neighbor decoding agree when the geometric advantage from candidate-distance separation exceeds the base-score difference. This condition connects the distance-separation and decoding-agreement diagnostics in Section~\ref{sec:main:trajectory}.

Probability concentration also controls the nearest neighbor of the continuous prediction. Write the probabilities at this position as $p$ and the continuous prediction as $m=E^\top p$. Using minimum separation $\Delta$ and diameter $D_E$ from Eq.~\eqref{eq:app:C.25},
\begin{equation}
\|m-e_j\|=\left\|\sum_{v\ne j}p_v(e_v-e_j)\right\|
\le D_E(1-p_j).
\tag{C.36}\label{eq:app:C.37}
\end{equation}
Thus, if $D_E(1-p_j)<\Delta/2$, the unique nearest embedding to $m$ is also $e_j$: for every $v\ne j$, the triangle inequality gives $\|m-e_v\|\ge\Delta-\|m-e_j\|>\|m-e_j\|$. Together with the distance-gap condition in Eq.~\eqref{eq:app:C.36}, this ensures that probability argmax, state nearest-neighbor decoding, and readout nearest-neighbor decoding all select token $j$.

\section{Training and Sampling Implementation}\label{sec:app:D}

\subsection{Data Splits}\label{sec:app:D.data}

Preprocessed OWT sequences are partitioned in their stored order: the first 90\% for training, 90\%--95\% for calibration, 95\%--97.5\% for validation, and the final 2.5\% for testing. We exclude 64 rows on each side of every internal split boundary. LM1B retains its official training and test splits. Within each dataset, all methods and training seeds use the same splits.

\subsection{Optimization Settings and Parameter Updates}\label{sec:app:D.2}

$\operatorname{sg}$ denotes stop-gradient.

Both datasets use one gradient-accumulation step. The task and schedule optimizers are separate AdamW instances with betas $(0.9,0.999)$, epsilon $10^{-8}$, and zero weight decay. For zero-based update index $k$, both use learning rate $3\times10^{-4}\min\{1,(k+1)/2500\}$. Task gradients are clipped at norm 1; parameters and Adam states are stored in FP32.

Continuous MSE is averaged over token positions and embedding coordinates and uses the noise weight below; CE is averaged over token positions without additional weighting.

\textbf{Primary-task and schedule updates.} To weight task contributions across noise levels, we use the positive SNR-based weight

\begin{equation}
w(\gamma)=\left(1+\frac{\alpha_\gamma}{\sigma_\gamma}\right)^2
=(1+e^{-\gamma/2})^2.
\tag{D.1}\label{eq:app:D.1}
\end{equation}

The weight $w$ sets the contribution of errors at each noise level, while $\pi_\psi$ determines how often each noise level is sampled. They enter expected-loss weighting and sampling, respectively.

The task optimizer updates $\theta$ with frozen embeddings and $(\theta,A)$ in the joint setting. The schedule is then updated from stop-gradient measurements collected during the same task forward passes, as specified in Algorithm~D1.

\textbf{Observation-branch warmup.} Equation~\eqref{eq:main:15} adds network and Gaussian observation scores. Early in training, the observation branch is gradually enabled. For warmup length $K_{\mathrm{skip}}=5000$, use

\begin{equation}
r_k=\min(1,k/K_{\mathrm{skip}}).
\tag{D.2}\label{eq:app:D.2}
\end{equation}

At $r_k=0$ only network scores are used; at $r_k=1$ the full Gaussian branch is active. With equal-norm embeddings, warmup probabilities satisfy $p_{\ell,v}\propto\omega_{\ell,v}K_{\ell,v}^{r_k}$.

\textbf{Parameter averaging.} After task and schedule updates, average network and schedule parameters. Treat raw embedding parameters $A$ according to whether they are frozen:

\begin{equation}
\begin{aligned}
(\bar\theta,\bar\psi)&\leftarrow\tau(\bar\theta,\bar\psi)+(1-\tau)(\theta,\psi),\\
\bar A&\leftarrow
\begin{cases}
A,&\text{frozen embeddings},\\
\tau\bar A+(1-\tau)A,&\text{joint updates},
\end{cases}\\
E_{\mathrm{EMA}}&=\operatorname{RowNorm}(\bar A).
\end{aligned}
\tag{D.3}\label{eq:app:D.3}
\end{equation}

Here $\tau\in[0,1)$ is the EMA decay, bars denote averaged parameters, and EMA is initialized by copying the initial model. Joint settings average raw $A$ before RowNorm constructs $E$; frozen settings copy pretrained $A$ directly. Evaluation uses one consistent EMA backbone, vocabulary head, embedding matrix, and schedule.

\subsection{Gumbel Fitting and Independent Schedule Updates}\label{sec:app:D.3}

TIBF and the continuous-MSE control fit the information-change measurement $\widehat g=e^{-\gamma}\|X-M\|_F^2/(2L)$ in Eq.~\eqref{eq:app:A.11} with a Gumbel density, where $M=PE$ is the pre-update continuous prediction. Let $\mu\in\mathbb R$ control location, $\beta>0$ control scale, and $C>0$ match the measurement amplitude, with $\psi=(\mu,\beta,C)$. Denote the CDF and density by $F_\psi,f_\psi$; $C$ affects regression amplitude but not distribution shape. The CDF is $F_\psi(\gamma)=\exp[-\exp(- (\gamma-\mu)/\beta)]$, whose derivative gives the density used in the squared-error fitting objective

\begin{equation}
\begin{aligned}
f_\psi(\gamma)&=\frac1\beta
\exp\!\left[-\frac{\gamma-\mu}{\beta}
-\exp\!\left(-\frac{\gamma-\mu}{\beta}\right)\right],\\
\mathcal L_{\mathrm{sched}}(\psi)&=
\mathbb E\!\left[
\left(Cf_\psi(\gamma)
-\operatorname{sg}(\widehat g)\right)^2\right].
\end{aligned}
\tag{D.4}\label{eq:app:D.4}
\end{equation}

The expectation is estimated by the current batch average. Sampled $\gamma$ and measured $\widehat g$ are treated as fixed inputs, so fitting updates only $\psi$. After each update, $\beta$ and $C$ are constrained to be at least $10^{-4}$ to maintain positive scale and amplitude.

\textbf{Training and generation.} To avoid extreme distribution tails, choose truncation probability $0<\varepsilon_q<1/2$, set to $10^{-5}$ in our experiments. The inverse CDF $F_\psi^{-1}$ determines the window and actual sampling density:

\begin{equation}
\begin{aligned}
\gamma_{\mathrm{lo}}&=F_\psi^{-1}(\varepsilon_q),
&\gamma_{\mathrm{hi}}&=F_\psi^{-1}(1-\varepsilon_q),\\
\pi_\psi(\gamma)&=\frac{f_\psi(\gamma)}{1-2\varepsilon_q}
\mathbf1_{[\gamma_{\mathrm{lo}},\gamma_{\mathrm{hi}}]}(\gamma).
\end{aligned}
\tag{D.5}\label{eq:app:D.5}
\end{equation}

Here $\mathbf1$ is an interval indicator, and the denominator normalizes the truncated density. Training samples quantiles in the truncated probability interval and maps them to $\gamma$ through the inverse CDF. Generation uses the final EMA schedule with uniformly spaced decreasing quantiles.

Setting $u=F_\psi(\gamma)$ and taking logarithms twice gives $\log(-\log u)=-(\gamma-\mu)/\beta$. Solving for $\gamma$ yields
\begin{equation}
F_\psi^{-1}(u)=\mu-\beta\log(-\log u),\qquad 0<u<1.
\tag{D.6}\label{eq:app:D.6}
\end{equation}

The truncated quantile interval retains probability mass $1-2\varepsilon_q$, so its CDF is $(F_\psi(\gamma)-\varepsilon_q)/(1-2\varepsilon_q)$. If $g=Cf_\psi$ throughout the window, total information change is $\Delta\mathsf H=C(1-2\varepsilon_q)$, and Eq.~\eqref{eq:app:D.5} recovers the ideal sampling density in Eq.~\eqref{eq:app:A.10}. Equation~\eqref{eq:app:D.4} tracks the information-change rate with a Gumbel approximation.

\textbf{CDF fitting for the CE controls.} The LangFlow controls use the per-sequence mean cross-entropy measured before the task update, $\widehat c=-L^{-1}\sum_{\ell=1}^L\log p_{\ell,Y_\ell}$, as the schedule target. Using the Gumbel CDF $F_\psi$ defined above, a positive amplitude $C$ scales the fitted curve:
\begin{equation}
\mathcal L_{\mathrm{sched}}^{\mathrm{CE}}(\psi)
=\mathbb E\!\left[\left(CF_\psi(\gamma)-\operatorname{sg}(\widehat c)\right)^2\right].
\tag{D.7}
\label{eq:app:D.added19}
\end{equation}
Here $C$ controls the amplitude of the CE curve. Both schedules follow the information-uniform principle. At a given noise level, when the predictor matches the true posterior, the expected CE equals the average of the position-wise conditional token entropies~\citep{chen2026langflow}. Equation~\eqref{eq:app:D.4} uses denoising error to estimate how the joint conditional entropy of the sequence, normalized by sequence length, changes with noise level. The sampled $\gamma$ and measured $\widehat c$ are held fixed; network and embedding parameters are frozen while only schedule parameters are updated. The separate optimizer, positivity constraints, truncated-quantile sampling, and generation grid follow the same settings as above.

\subsection{Euler Sampling and Forward-Pass Budget}\label{sec:app:D.4}

\textbf{Generation grid.} Fix the final EMA parameters, disable dropout, and fully enable the Gaussian branch. For total forward-pass budget $N\ge2$, final Gumbel parameters $\mu,\beta$, and quantile truncation $\varepsilon_q$, set

\begin{equation}
u_i=1-\varepsilon_q-\frac{i}{N-1}(1-2\varepsilon_q),
\qquad \gamma_i=F_\psi^{-1}(u_i),
\quad i=0,\ldots,N-1.
\tag{D.8}\label{eq:app:D.7}
\end{equation}

Here $i$ indexes nodes, $u_i$ are decreasing quantiles, and $\gamma_i$ are the corresponding noise levels. Quantiles are equally spaced, while noise levels run from high to low.

Let $Z_i\in\mathbb R^{L\times d}$ denote the continuous state at $\gamma_i$. Initialize $Z_0$ with independent standard Gaussian entries and set the first self-conditioning input to $S_0=0$. Denote the sequence probabilities in Eq.~\eqref{eq:main:15}, with self-conditioning input $S$, by $P_\theta(Z,\gamma;S)$. At each node, compute $P_i=P_\theta(Z_i,\gamma_i;S_i)$ and $M_i=P_iE$ to update the state.

\textbf{State updates.} For the velocity in Eq.~\eqref{eq:main:18}, $v_\theta=(\alpha_\gamma^2Z_\gamma-\alpha_\gamma M)/2$, define $\widetilde Z_\gamma=Z_\gamma/\sigma_\gamma$ and $\eta_\gamma=\alpha_\gamma/\sigma_\gamma=e^{-\gamma/2}$. Differentiating the rescaled state cancels the scaling terms, giving

\begin{equation}
\begin{aligned}
\frac{\mathrm d}{\mathrm d\gamma}\left(\frac{Z_\gamma}{\sigma_\gamma}\right)
&=\frac{v_\theta}{\sigma_\gamma}
-\frac{\sigma_\gamma'}{\sigma_\gamma^2}Z_\gamma\\
&=\left(\frac{\alpha_\gamma^2}{2\sigma_\gamma}
-\frac{\sigma_\gamma'}{\sigma_\gamma^2}\right)Z_\gamma
-\frac{\alpha_\gamma}{2\sigma_\gamma}M\\
&=-\frac{\eta_\gamma}{2}M
=\frac{\mathrm d\eta_\gamma}{\mathrm d\gamma}M.
\end{aligned}
\tag{D.9}
\label{eq:app:D.added20}
\end{equation}

Here $\sigma_\gamma'=\sigma_\gamma\alpha_\gamma^2/2$ and $\eta_\gamma'=-\eta_\gamma/2$. Thus $\mathrm d\widetilde Z/\mathrm d\eta=M$: in $\eta$ coordinates, the continuous prediction $M$ is the direction of motion of the scaled state. Write $\alpha_i=\alpha_{\gamma_i}$, $\sigma_i=\sigma_{\gamma_i}$, $\eta_i=\alpha_i/\sigma_i$, and $\widetilde Z_i=Z_i/\sigma_i$. Decreasing $\gamma$ increases $\eta$, so a generation step has $\eta_{i+1}>\eta_i$. Holding $M_i$ fixed within interval $i$ gives a first-order Euler step:

\begin{equation}
\widetilde Z_{i+1}
=\widetilde Z_i+(\eta_{i+1}-\eta_i)M_i.
\tag{D.10}
\label{eq:app:D.added21}
\end{equation}

Multiplying by the next noise scale $\sigma_{i+1}$ restores the original state and expresses the same update using predicted noise:

\begin{equation}
\begin{aligned}
\widehat\varepsilon_i&=\frac{Z_i-\alpha_iM_i}{\sigma_i},\\
Z_{i+1}
&=\sigma_{i+1}\left[\frac{Z_i}{\sigma_i}
+\left(\frac{\alpha_{i+1}}{\sigma_{i+1}}
-\frac{\alpha_i}{\sigma_i}\right)M_i\right]\\
&=\alpha_{i+1}M_i+\sigma_{i+1}\widehat\varepsilon_i,\\
S_{i+1}&=M_i.
\end{aligned}
\tag{D.11}\label{eq:app:D.8}
\end{equation}

Here $i=0,\ldots,N-2$. The direction $\widehat\varepsilon_i$ is inferred from the current state and clean prediction and is reused without drawing independent noise. Cache $S_{i+1}$ stores the current prediction for the next node. Both $P_{i+1}$ and $M_{i+1}$ are recomputed after each update; predictions are held constant only within an interval.

\textbf{Final decoding and call counts.} There are $N-1$ state intervals between $N$ grid points. The first $N-1$ network calls each provide probabilities and readouts for one update. At $Z_{N-1}$, a new prediction at the final noise level is required for decoding. Set

\begin{equation}
\widehat Y_\ell=\operatorname*{arg\,max}_{v\in[V]}
 s_{\ell,v}(Z_{N-1},\gamma_{N-1};S_{N-1}),
\qquad \mathrm{NFE}=(N-1)+1=N.
\tag{D.12}\label{eq:app:D.9}
\end{equation}

Softmax preserves candidate ordering, so argmax over logits and probabilities gives the same token. The budget includes final decoding; the primary settings use $N=256$ on OWT and $N=128$ on LM1B.

\subsection{Training and Sampling Algorithms}\label{sec:app:D.5}

$X(Y)$ denotes the embedding map in Eq.~\eqref{eq:main:4}, evaluated by looking up token embeddings in the current $E$. Forward returns the total score matrix $s\in\mathbb R^{L\times V}$ in Eq.~\eqref{eq:main:15}; its final argument is the observation-branch strength. Readout computes $\operatorname{softmax}(s)E$ in position chunks, and None denotes a zero self-conditioning input. ChunkedBrier returns the weighted mean Brier loss for one microbatch and a detached continuous readout $M$. The loss is normalized by $BL$, with backward computation as in Appendix~\ref{sec:app:D.6}. RowNorm is defined in Eq.~\eqref{eq:main:11}, and $F_\psi^{-1}$ in Eq.~\eqref{eq:app:D.6}. Quantiles are denoted by $u_i,u_j$, with independent within-stratum uniforms $\upsilon_j$.

\textbf{Algorithm D1: TIBF training.} Inputs are $P_{\mathrm{data}}$, initial parameters $(\theta,A,\psi)$, total updates $K_{\mathrm{train}}$, microbatch size $B$, accumulation count $G$, self-conditioning probability $p_{\mathrm{SC}}$, EMA decay $\tau$, quantile truncation $\varepsilon_q$, and the optimization settings above. Outputs are final parameters and their EMA counterparts.

\begin{Verbatim}[breaklines=true,breakanywhere=true,fontsize=\footnotesize]
Disable outer autocast; enable BF16 only inside the DiT
Opt_main <- AdamW(theta) if E is frozen, else AdamW(theta, A)
Opt_sched <- AdamW(psi)
(theta_ema, A_ema, psi_ema) <- copy(theta, A, psi)
for k = 0, ..., K_train - 1 do
    Enable training mode and dropout
    Set both learning rates and branch strength r_k (Eq. D.2)
    Set requires_grad(theta) = True and requires_grad(A) = not frozen
    Freeze psi and clear its gradients; Opt_main.zero_grad()

    for j = 0, ..., BG - 1 do
        upsilon_j ~ Uniform(0, 1)
        u_j <- eps_q + (1 - 2eps_q)(j + upsilon_j)/(BG)
    end for
    Shuffle {u_j}; Gamma <- sg(F^(-1)_{mu,beta}({u_j}))
    Buffer <- empty list

    for g = 1, ..., G do
        gamma <- g-th B-element slice of Gamma
        Sample B length-L token sequences Y
        eps <- standard Gaussian noise of shape B x L x d
        E <- RowNorm(A); X <- X(Y)
        Z <- alpha_gamma X + sigma_gamma eps
        active <- B independent Bernoulli(p_SC) draws
        S <- None
        if any(active) then
            with no_grad and autocast weight caching disabled:
                s0 <- Forward(theta, E, Z[active], gamma[active], None; r_k)
                S <- zeros_like(Z)
                S[active] <- Readout(s0, E)
        end if

        s <- Forward(theta, E, Z, gamma, S; r_k)
        loss, M <- ChunkedBrier(s, Y, w(gamma), sg(E))
        with no_grad:
            for b = 1, ..., B do
                e_hat_b <- ||X_b - M_b||_F^2 / (Ld)
                g_hat_b <- (d/2) exp(-gamma_b) e_hat_b
                Append sg(gamma_b, g_hat_b) to Buffer
            end for
        Backpropagate loss / G to primary optimizer parameters
    end for

    Clip the joint norm of primary gradients to 1; Opt_main.step()
    Freeze theta and A; clear their gradients
    Enable psi gradients; Opt_sched.zero_grad()
    L_sched <- mean([C f_{mu,beta}(gamma_b) - g_hat_b]^2 over Buffer)
    Backpropagate L_sched only to psi; Opt_sched.step()
    beta <- max(beta, 1e-4); C <- max(C, 1e-4)
    (theta_ema, psi_ema) <- tau(theta_ema, psi_ema) + (1 - tau)(theta, psi)
    A_ema <- A if E is frozen, else tau A_ema + (1 - tau) A
    Save a full checkpoint if scheduled; then validate EMA if scheduled
end for
return (theta, A, psi), (theta_ema, A_ema, psi_ema)
\end{Verbatim}

\textbf{Algorithm D2: TIBF continuous sampling.} Inputs are complete EMA parameters $(\bar\theta,\bar A,\bar\psi)$, length $L$, total evaluation budget $N\ge2$, and quantile truncation $\varepsilon_q$. We write the algorithm for one sequence; multiple sequences can be processed in parallel. Dropout and gradient recording are disabled, and observation-branch strength is one.

\begin{Verbatim}[breaklines=true,breakanywhere=true,fontsize=\footnotesize]
Disable dropout, gradient recording, and outer autocast
E <- RowNorm(A_ema); read mu, beta from psi_ema
for i = 0, ..., N - 1 do
    u_i <- FP32(1 - eps_q - i(1 - 2eps_q)/(N - 1))
    gamma_i <- FP32(F^(-1)_{mu,beta}(u_i))
end for
Z <- FP32(standard Gaussian noise of shape L x d)
S <- None; calls <- 0
for i = 0, ..., N - 2 do
    s <- Forward(theta_ema, E, Z, gamma_i, S; 1)
    M <- Readout(s, E); calls <- calls + 1
    Z <- EulerDataStep(Z, M, gamma_i, gamma_{i+1})
    S <- M
end for
s_final <- Forward(theta_ema, E, Z, gamma_{N-1}, S; 1)
calls <- calls + 1
Y_hat_ell <- argmax_v s_final[ell, v] for each position ell
assert calls = N
return Y_hat

procedure EulerDataStep(Z, M, gamma_cur, gamma_next):
    Convert gamma_cur, gamma_next, Z, M to FP64
    cur <- Z * exp((softplus(gamma_cur) - softplus(gamma_next))/2)
    end <- sqrt(sigmoid(-gamma_next)) * M
    return FP32(end + exp((gamma_next - gamma_cur)/2) * (cur - end))
\end{Verbatim}

Here $\operatorname{softplus}(x)=\log(1+e^x)$. Probabilities, grid nodes, and stored states use FP32; coefficients and intermediate state combinations use FP64. Since $\alpha_\gamma=(1+e^\gamma)^{-1/2}$, the first exponential factor in Algorithm~D2 equals $\alpha_{i+1}/\alpha_i$. Using $\exp[(\gamma_{i+1}-\gamma_i)/2]=\sigma_{i+1}\alpha_i/(\sigma_i\alpha_{i+1})$ then recovers Eq.~\eqref{eq:app:D.8}.

\subsection{Chunked Brier Loss Computation and Numerical Precision}\label{sec:app:D.6}

The loss at each position is $\frac12[(1-p_y)^2+\sum_{v\ne y}p_v^2]$. The implementation computes full-vocabulary softmax, squared sums, and loss accumulation in FP32, chunking over positions with a default chunk size of 256 and no vocabulary truncation. The chunked softmax is recomputed during backpropagation. The loss gradient with respect to $p$ is $p-\delta_y$, and the softmax Jacobian is $\operatorname{diag}(p)-pp^\top$, where $\operatorname{diag}(p)$ places $p$ on the diagonal. The chain rule gives
\begin{equation}
\nabla_s\mathcal L=(\operatorname{diag}(p)-pp^\top)(p-\delta_y),\qquad p=\operatorname{softmax}(s).
\tag{D.13}\label{eq:app:D.backward}
\end{equation}
Multiplying each position gradient by $w(\gamma)/(GBL)$ gives its contribution during gradient accumulation. Only the chunked softmax is recomputed; the backbone forward pass is reused.

\textbf{Numerical precision.} The unified implementation disables outer autocast during training, training-time validation, and default generation. BF16 mixed precision is enabled only within DiT, where LayerNorm still uses FP32. The time-conditioning MLP, self-conditioning projection, and probability-weighted embedding readout $PE$ use FP32. Gaussian dot products use FP32 and are cast to the backbone logit dtype before scores are combined; full-vocabulary softmax and Brier computations follow the FP32 path described above. Continuous states are stored in FP32; only Euler coefficients and intermediate state combinations use FP64. Training and default generation allow TF32; distance searches and low-noise diagnostics disable it.

\subsection{Self-Conditioning during Training and Generation}\label{sec:app:D.7}

Self-conditioning uses the continuous readout in Eq.~\eqref{eq:main:17} as auxiliary input.

\textbf{Same-point self-conditioning during training.} Independently draw $m_b\sim\operatorname{Bernoulli}(p_{\mathrm{SC}})$ for each sequence $b$, with default activation probability $p_{\mathrm{SC}}=0.25$. Self-conditioning is enabled when $m_b=1$. Let $Y_b$ be sequence $b$ and $Z_b=\alpha_{\gamma_b}X(Y_b)+\sigma_{\gamma_b}\varepsilon_b$ its noisy input, with independent $\varepsilon_b$ as in Eq.~\eqref{eq:main:5}. When enabled, first compute at the same input

\begin{equation}
P_b^{(0)}=P_\theta(Z_b,\gamma_b;0),
\qquad
S_b=\operatorname{sg}(P_b^{(0)}E).
\tag{D.14}\label{eq:app:D.10}
\end{equation}

Superscript $(0)$ denotes the preliminary prediction without auxiliary input. When self-conditioning is disabled, set $S_b=0$ without an auxiliary pass. The main pass then uses $S_b$ to compute the task loss. Stop-gradient covers the entire auxiliary branch, including its network, noisy input, and embedding readout; joint settings retain gradients to $E$ through the main pass. Autocast weight caching is disabled during the auxiliary pass to preserve gradient propagation through trainable weights in the main pass. Both passes use the same noisy observation.

Generation uses the cross-step cache in Eq.~\eqref{eq:app:D.8} and Algorithm~D2, whereas training uses the stop-gradient same-point prediction above.

\section{Generative Scoring and Dataset PPL Evaluation Protocols}\label{sec:app:E}

\subsection{Gen.PPL Scoring and Statistical Methods}\label{sec:app:E.1}

Generation experiments use the final 100,000-step EMA checkpoints from training seeds 17, 42, and 1234, with GPU seed 42 for generation. Within each corpus, primary evaluation, sampling-budget sweeps, temperature studies, and paired comparisons share initial Gaussian noise and sample indices. Each checkpoint generates 1,024 sequences: length 1,024 and batch size 8 on OWT, and length 128 and batch size 32 on LM1B. Sampling grids and forward-pass counts follow Appendix~\ref{sec:app:D.4} and Algorithm~D2.

Both corpora use the same GPT-2 large scoring procedure. Samples are decoded with their respective generator tokenizers, retaining literal special tokens, and retokenized by the scorer. Let $S_{\mathrm{gen}}$ be the sample count, $\widetilde Y_{\mathrm{ref}}^{(i)}$ the sequence actually received by the scorer, $T_i\ge2$ its unpadded length after truncation, and $p_{\mathrm{ref}}$ the scoring language model. All logarithms below are natural. Aggregation over valid next-token predictions gives
\begin{equation}
\begin{aligned}
\mathrm{NLL}_{\mathrm{gen}}
&=-\frac{\displaystyle\sum_{i=1}^{S_{\mathrm{gen}}}\sum_{\ell=2}^{T_i}
\log p_{\mathrm{ref}}(\widetilde Y_{\mathrm{ref},\ell}^{(i)}\mid\widetilde Y_{\mathrm{ref},<\ell}^{(i)})}
{\displaystyle\sum_{i=1}^{S_{\mathrm{gen}}}(T_i-1)},\\
\mathrm{Gen.PPL}&=\exp(\mathrm{NLL}_{\mathrm{gen}}).
\end{aligned}
\tag{E.1}\label{eq:app:E.score}
\end{equation}

Both corpora use FP16 and truncation to at most 1,024 tokens. Scoring batch sizes are 4 on OWT and 32 on LM1B. Batches are right-padded, with padded labels ignored. Each batch mean loss is multiplied by its valid next-token count before summation and normalization by the total count.

Results report arithmetic means and sample standard deviations across the three training seeds. Paired bootstrap comparisons of mean Gen.PPL differences jointly resample sequence indices associated with common initial noise; see Appendix~\ref{sec:app:F.paired}.

\subsection{ODE Evaluation Model and Variational Bound}\label{sec:app:E.2}

Fix the final model parameters and embeddings $E$. Dataset PPL recomputes same-point self-conditioning at every state, defining the evaluation field
\begin{equation}
S^{(0)}(Z,\gamma)=P_\theta(Z,\gamma;0)E,\qquad
M_\theta(Z,\gamma)=P_\theta(Z,\gamma;S^{(0)}(Z,\gamma))E.
\tag{E.2}
\label{eq:app:E.added22}
\end{equation}
Write the second prediction's probability for token $v$ at position $\ell$ as $p_{\theta,\ell}(v\mid Z,\gamma)$.

Choose finite noise endpoints $a<b$ and retain $\eta=e^{-\gamma/2}$ and $\widetilde Z=Z/\sigma_\gamma$, with $\eta_a=e^{-a/2}>\eta_b=e^{-b/2}$ and $\sigma_a=\sigma_{\gamma=a}$. Appendix~\ref{sec:app:D.4} gives the evaluation ODE $\mathrm d\widetilde Z/\mathrm d\eta=M_\theta(\sigma_\gamma\widetilde Z,\gamma)$. Assume this field is differentiable and locally Lipschitz in state. With fixed $E$, its readout lies in the convex hull of finitely many embeddings and is bounded, defining an invertible flow over the finite integration interval.

Let $I_{Ld}$ be the $Ld$-dimensional identity matrix. Assign a standard Gaussian prior to $\operatorname{vec}(\widetilde Z_b)$ at high noise, and denote the induced low-noise density by $p_a^{\mathrm{sc}}$. The low-noise encoder and categorical decoder are
\begin{equation}
\begin{aligned}
q_a^{\mathrm{sc}}(\widetilde Z\mid Y)
&=\mathcal N(\operatorname{vec}(\widetilde Z);\eta_a\operatorname{vec}(X(Y)),I_{Ld}),\\
p_{\mathrm{dec}}(Y\mid\widetilde Z)
&=\prod_{\ell=1}^L p_{\theta,\ell}(Y_\ell\mid\sigma_a\widetilde Z,a).
\end{aligned}
\tag{E.3}
\label{eq:app:E.added23}
\end{equation}
These distributions define a discrete sequence probability $p_{\mathrm{ODE}}$. Expressing its integral as an encoder expectation and applying Jensen's inequality gives
\begin{equation}
\begin{aligned}
p_{\mathrm{ODE}}(Y)
&=\int p_a^{\mathrm{sc}}(\widetilde Z)p_{\mathrm{dec}}(Y\mid\widetilde Z)\,\mathrm d\widetilde Z\\
&=\mathbb E_{q_a^{\mathrm{sc}}}\!\left[
\frac{p_a^{\mathrm{sc}}(\widetilde Z)p_{\mathrm{dec}}(Y\mid\widetilde Z)}{q_a^{\mathrm{sc}}(\widetilde Z\mid Y)}\right],\\
-\log p_{\mathrm{ODE}}(Y)
&\le\mathcal L_{\mathrm{NELBO}}(Y),\\
\mathcal L_{\mathrm{NELBO}}(Y)
&:=\mathbb E_{q_a^{\mathrm{sc}}}\!\left[
-\log p_{\mathrm{dec}}(Y\mid\widetilde Z)
+\log\frac{q_a^{\mathrm{sc}}(\widetilde Z\mid Y)}{p_a^{\mathrm{sc}}(\widetilde Z)}\right].
\end{aligned}
\tag{E.4}\label{eq:app:E.1}
\end{equation}

\subsection{Density Change of Variables and Divergence Estimation}\label{sec:app:E.3}

The chain rule for $Z=\sigma_\gamma\widetilde Z$ gives divergence $\sigma_\gamma\operatorname{div}_ZM_\theta$ with respect to the scaled state. Let $p_\eta^{\mathrm{sc}}$ denote the flow density at coordinate $\eta$. The continuity equation yields
\begin{equation}
\frac{\mathrm d}{\mathrm d\eta}\log p_\eta^{\mathrm{sc}}(\widetilde Z(\eta))
=-\sigma_\gamma\operatorname{div}_ZM_\theta(Z(\eta),\gamma),
\qquad \gamma=-2\log\eta.
\tag{E.5}
\label{eq:app:E.added24}
\end{equation}
Let $\widetilde Z_a$ and $\widetilde Z_b$ denote states on the same trajectory at $\eta_a$ and $\eta_b$. Integrating along this trajectory from $\eta_a$ to $\eta_b$, define
$I_{\mathrm{div}}=\int_{\eta_a}^{\eta_b}\sigma_\gamma\operatorname{div}_ZM_\theta(Z(\eta),\gamma)\,\mathrm d\eta$.
The high-noise prior and encoder entropy give
\begin{equation}
\begin{aligned}
\log p_a^{\mathrm{sc}}(\widetilde Z_a)
&=\log\mathcal N(\operatorname{vec}(\widetilde Z_b);0,I_{Ld})+I_{\mathrm{div}},\\
\mathbb E_{q_a^{\mathrm{sc}}}\log q_a^{\mathrm{sc}}(\widetilde Z_a\mid Y)
&=-\frac{Ld}{2}\log(2\pi e).
\end{aligned}
\tag{E.6}\label{eq:app:E.2}
\end{equation}
Expanding the Gaussian log density as $-Ld\log(2\pi)/2-\|\widetilde Z_b\|_F^2/2$ and substituting into Eq.~\eqref{eq:app:E.1} cancels the Gaussian constants, yielding
\begin{equation}
\begin{aligned}
\mathcal L_{\mathrm{NELBO}}(Y)
=\mathbb E_{\widetilde Z_a\sim q_a^{\mathrm{sc}}(\cdot\mid Y)}\!\Bigg[
&-\sum_{\ell=1}^L\log p_{\theta,\ell}(Y_\ell\mid\sigma_a\widetilde Z_a,a)\\
&-I_{\mathrm{div}}+\frac12\|\widetilde Z_b\|_F^2-\frac{Ld}{2}\Bigg].
\end{aligned}
\tag{E.7}\label{eq:app:E.3}
\end{equation}
The terms are the low-noise reconstruction NLL, negative divergence integral, and prior contribution after cancellation of the encoder entropy constant. One encoder-noise draw per sequence estimates this expectation.

Let $J_M=\partial\operatorname{vec}M_\theta/\partial\operatorname{vec}Z\in\mathbb R^{Ld\times Ld}$. At every field evaluation, draw an independent probe $\xi\sim\mathcal N(0,I_{Ld})$. Since $\mathbb E_\xi[\xi\xi^\top]=I_{Ld}$, divergence is estimated by
\begin{equation}
\widehat{\operatorname{div}_ZM_\theta}=\xi^\top J_M\xi,
\qquad
\mathbb E_\xi[\xi^\top J_M\xi]
=\operatorname{tr}\!\left(J_M\mathbb E_\xi[\xi\xi^\top]\right)
=\operatorname{tr}J_M.
\tag{E.8}\label{eq:app:E.4}
\end{equation}

Attached differentiation retains the full derivative of the same-point composite readout, as required by the change of variables above. Detached differentiation holds auxiliary input $S^{(0)}$ fixed and computes only the primary prediction's partial derivative with respect to $Z$. Both protocols recompute forward self-conditioning; they differ in the differentiation path used for divergence.

\subsection{Numerical Protocol for Dataset PPL}\label{sec:app:E.4}

OWT and LM1B use the same probability-flow ODE evaluation framework~\citep{chen2026langflow}, reporting attached and detached scores separately. Both compute the reconstruction, negative divergence integral, and prior contributions in Eq.~\eqref{eq:app:E.3}, estimating divergence along their respective differentiation paths. Let $\widehat{\mathcal J}(Y^{(i)})$ be the sum of these contributions for held-out sequence $i$, $L_i$ its valid-token count, and $S_{\mathrm{eval}}$ the sequence count. Token-weighted aggregation gives
\begin{equation}
\widehat{\mathrm{PPL}}
=\exp\!\left[
\frac{\sum_{i=1}^{S_{\mathrm{eval}}}\widehat{\mathcal J}(Y^{(i)})}
{\sum_{i=1}^{S_{\mathrm{eval}}}L_i}\right].
\tag{E.9}\label{eq:app:E.aggregate}
\end{equation}

Both corpora and both protocols use FP32 states, BF16 mixed precision within DiT, FP32 Gaussian dot products, and TF32-enabled matrix multiplications. Integration uses 128 Heun-2 steps in $(\widetilde Z,\eta)$ coordinates. Its 129 nodes start from equally spaced probabilities in $[0,1]$, clipped to $[10^{-5},1-10^{-5}]$ and mapped through the final EMA schedule's inverse CDF to $\gamma$, then to $\eta=e^{-\gamma/2}$. Endpoints $a,b$ are the minimum and maximum grid noise levels. The evaluation RNG seed is 1.

Attached evaluation uses batch size 2 on both corpora; detached evaluation uses batch sizes 2 and 16 on OWT and LM1B, respectively. Zero-shot transfer uses the detached protocol.

OWT uses validation data and LM1B the official test split, each with 1,024 fixed sequences of lengths 1,024 and 128, respectively. Zero-shot transfer uses OWT models at length 1,024 and at most 1,024 sequences per target corpus. Three-seed summaries follow the statistical procedure in Appendix~\ref{sec:app:E.1}.

\section{Supplementary Generation and Transfer Experiments}\label{sec:app:F}

This appendix reports generation quality and diversity, diversity across sampling budgets, temperature sensitivity, paired differences between models, and zero-shot transfer. Common generation and scoring settings are given in Appendix~\ref{sec:app:E.1}.

\subsection{Diversity Metrics}\label{sec:app:F.metrics}

Following diversity evaluation for continuous language generation~\citep{li2026convergeflow,chen2026langflow}, we compute token entropy from empirical token frequencies within each generated sequence. Retain the sample count $S_{\mathrm{gen}}$, and write sequence $i$ of length $L$ as $\widetilde Y^{(i)}$. Its empirical frequency for token $v$ is $\widehat p_v^{(i)}=L^{-1}\sum_{\ell=1}^L\mathbf1\{\widetilde Y_\ell^{(i)}=v\}$. Averaging sequence-level entropies gives

\begin{equation}
\widehat H_{\mathrm{tok}}
=-\frac1{S_{\mathrm{gen}}}\sum_{i=1}^{S_{\mathrm{gen}}}\sum_{v=1}^V
\widehat p_v^{(i)}\log\widehat p_v^{(i)},\qquad 0\log0=0.
\tag{F.1}\label{eq:app:F.1}
\end{equation}

Token entropy is measured in nats and describes dispersion of within-sequence token usage. 

For $n\in\{2,4\}$, let $\mathcal G_n^{(i)}$ be the set of distinct consecutive $n$-grams in sequence $i$. A length-$L$ sequence contains $L-3$ overlapping 4-gram windows. The fraction of windows that repeat a previously observed 4-gram is therefore
\begin{equation}
\widehat r_4=\frac1{S_{\mathrm{gen}}}\sum_{i=1}^{S_{\mathrm{gen}}}
\left(1-\frac{|\mathcal G_4^{(i)}|}{L-3}\right),\qquad L\ge4.
\tag{F.2}\label{eq:app:F.2}
\end{equation}
Similarly, a sequence contains $L-1$ adjacent token pairs. Distinct-2 divides the number of distinct pairs by this window count and averages over sequences:
\begin{equation}
\mathrm{Distinct}\text{-}2
=\frac1{S_{\mathrm{gen}}}\sum_{i=1}^{S_{\mathrm{gen}}}
\frac{|\mathcal G_2^{(i)}|}{L-1}.
\tag{F.3}
\label{eq:app:F.added25}
\end{equation}
All metrics use the generating model's token IDs and retain special tokens. We compute the repeated 4-gram fraction per sequence and report it as a percentage in all tables.

\subsection{Generation Quality and Diversity}\label{sec:app:F.quality}

\begin{figure}
    \centering
    \includegraphics[width=\linewidth]{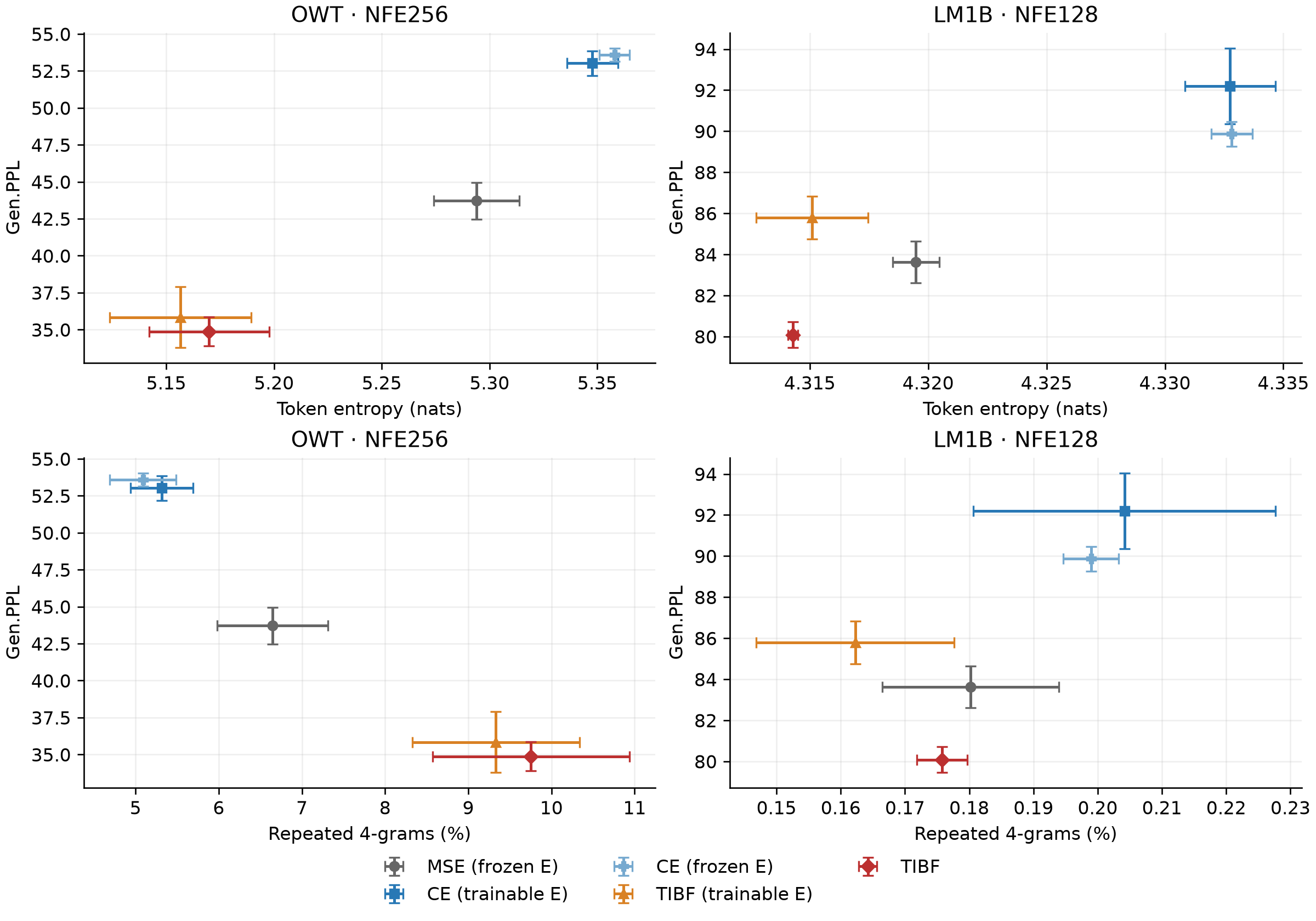}
    \caption{Generation quality and diversity under the common sampling settings. Left: OWT; right: LM1B. The top row plots Gen.PPL against token entropy; the bottom row plots Gen.PPL against repeated 4-gram fraction. Points are means across three training seeds; horizontal and vertical error bars show sample standard deviations, corresponding to Tables~\ref{tab5} and \ref{tab6}.}
    \label{fig3}
\end{figure}

Under the common primary evaluation settings, we examine quality--diversity trade-offs using token entropy and repeated 4-gram fraction, as shown in Tables~\ref{tab5} and \ref{tab6} and Figure~\ref{fig3}.

\begin{table}[htbp]
    \centering
    \caption{Generation quality and diversity on OWT. Mean $\pm$ sample SD across three training seeds.}
    \label{tab5}
    \begin{adjustbox}{max width=\linewidth}
\begin{tabular}{lccc}
\toprule
Model & Gen.PPL & Entropy & Repeated 4-grams (\%) \\ \midrule
ConvergeFlow & $43.71\pm1.23$ & $5.2939\pm0.0198$ & $6.6483\pm0.6661$ \\
LangFlow (trainable $E$) & $53.03\pm0.84$ & $5.3477\pm0.0119$ & $5.3179\pm0.3760$ \\
LangFlow (frozen $E$) & $53.59\pm0.46$ & $5.3580\pm0.0070$ & $\boldsymbol{5.0908\pm0.3990}$ \\
TIBF (trainable $E$) & $35.84\pm2.07$ & $5.1566\pm0.0327$ & $9.3303\pm1.0059$ \\
TIBF & $\boldsymbol{34.87\pm0.98}$ & $5.1699\pm0.0279$ & $9.7535\pm1.1837$ \\
\bottomrule\end{tabular}
\end{adjustbox}
\end{table}

\begin{table}[htbp]
    \centering
    \caption{Generation quality and diversity on LM1B. Mean $\pm$ sample SD across three training seeds.}
    \label{tab6}
    \begin{adjustbox}{max width=\linewidth}
\begin{tabular}{lccc}
\toprule
Model & Gen.PPL & Entropy & Repeated 4-grams (\%) \\ \midrule
ConvergeFlow & $83.63\pm1.01$ & $4.3195\pm0.0010$ & $0.1802\pm0.0138$ \\
LangFlow (trainable $E$) & $92.21\pm1.84$ & $4.3328\pm0.0019$ & $0.2042\pm0.0235$ \\
LangFlow (frozen $E$) & $89.87\pm0.60$ & $4.3328\pm0.0009$ & $0.1990\pm0.0043$ \\
TIBF (trainable $E$) & $85.79\pm1.03$ & $4.3151\pm0.0024$ & $\boldsymbol{0.1622\pm0.0154}$ \\
TIBF & $\boldsymbol{80.08\pm0.63}$ & $4.3143\pm0.0002$ & $0.1758\pm0.0039$ \\
\bottomrule\end{tabular}
\end{adjustbox}
\end{table}

On LM1B, TIBF retains the lowest mean Gen.PPL, with repeated 4-grams of $0.1758\pm0.0039\%$, lower than MSE and both CE controls; TIBF (trainable $E$) has slightly lower repetition. On OWT, TIBF has higher repetition ($9.75\pm1.18\%$ versus $6.65\pm0.67\%$ for MSE) and lower token entropy, reflecting a quality--diversity trade-off.

\FloatBarrier
\subsection{Supplementary Diversity across Sampling Budgets}\label{sec:app:F.budget}
The Gen.PPL comparisons appear in Section~\ref{sec:main:budget}. Tables~\ref{tab:owt_budget_diversity} and \ref{tab:lm1b_budget_diversity} report token entropy and repeated 4-gram percentages on OWT and LM1B. Figure~\ref{fig:owt_budget_entropy} combines these with the corresponding Gen.PPL values. Increasing NFE lowers mean Gen.PPL for every model on both corpora, while token entropy decreases and repetition increases. Temperature is fixed at $T=1$.

\FloatBarrier
\FloatBarrier
\refstepcounter{table}\label{tab:owt_budget_diversity}
\textbf{Table \thetable: OWT diversity across sampling budgets. Mean $\pm$ sample SD. Repeated 4-grams are percentages.}

\par\smallskip\noindent\begin{minipage}{\linewidth}
\textbf{Token entropy (nats)}\par\smallskip
\begingroup
\setlength{\tabcolsep}{2pt}
\scriptsize
\begin{adjustbox}{max width=\linewidth}
\begin{tabular}{lccccc}
\toprule
Model & NFE=64 & NFE=128 & NFE=256 & NFE=512 & NFE=1024 \\
\midrule
ConvergeFlow & $5.4497\pm0.0119$ & $5.3710\pm0.0142$ & $5.2939\pm0.0198$ & $5.2230\pm0.0243$ & $5.1583\pm0.0332$ \\
LangFlow (trainable $E$) & $5.4867\pm0.0041$ & $5.4134\pm0.0076$ & $5.3477\pm0.0119$ & $5.2898\pm0.0205$ & $5.2385\pm0.0246$ \\
LangFlow (frozen $E$) & $5.4899\pm0.0038$ & $5.4203\pm0.0050$ & $5.3580\pm0.0070$ & $5.3018\pm0.0109$ & $5.2542\pm0.0217$ \\
TIBF (trainable $E$) & $5.3683\pm0.0193$ & $5.2628\pm0.0252$ & $5.1566\pm0.0327$ & $5.0416\pm0.0347$ & $4.9367\pm0.0440$ \\
TIBF & $5.3801\pm0.0074$ & $5.2777\pm0.0207$ & $5.1699\pm0.0279$ & $5.0560\pm0.0330$ & $4.9491\pm0.0329$ \\
\bottomrule
\end{tabular}
\end{adjustbox}
\endgroup
\end{minipage}\par\smallskip

\par\smallskip\noindent\begin{minipage}{\linewidth}
\textbf{Repeated 4-grams (\%)}\par\smallskip
\begingroup
\setlength{\tabcolsep}{2pt}
\scriptsize
\begin{adjustbox}{max width=\linewidth}
\begin{tabular}{lccccc}
\toprule
Model & NFE=64 & NFE=128 & NFE=256 & NFE=512 & NFE=1024 \\
\midrule
ConvergeFlow & $2.88\pm0.24$ & $4.45\pm0.41$ & $6.65\pm0.67$ & $9.16\pm1.15$ & $11.93\pm1.71$ \\
LangFlow (trainable $E$) & $2.55\pm0.17$ & $3.83\pm0.31$ & $5.32\pm0.38$ & $7.03\pm0.70$ & $8.76\pm0.72$ \\
LangFlow (frozen $E$) & $\boldsymbol{2.48\pm0.10}$ & $\boldsymbol{3.67\pm0.18}$ & $\boldsymbol{5.09\pm0.40}$ & $\boldsymbol{6.69\pm0.57}$ & $\boldsymbol{8.24\pm0.82}$ \\
TIBF (trainable $E$) & $3.68\pm0.30$ & $5.94\pm0.60$ & $9.33\pm1.01$ & $13.75\pm1.41$ & $18.42\pm1.74$ \\
TIBF & $4.48\pm1.00$ & $6.58\pm1.00$ & $9.75\pm1.18$ & $13.81\pm1.35$ & $18.04\pm1.50$ \\
\bottomrule
\end{tabular}
\end{adjustbox}
\endgroup
\end{minipage}\par\smallskip

\FloatBarrier
\refstepcounter{table}\label{tab:lm1b_budget_diversity}
\textbf{Table \thetable: LM1B diversity across sampling budgets. Mean $\pm$ sample SD. Repeated 4-grams are percentages.}

\par\smallskip\noindent\begin{minipage}{\linewidth}
\textbf{Token entropy (nats)}\par\smallskip
\begingroup
\setlength{\tabcolsep}{2pt}
\scriptsize
\begin{adjustbox}{max width=\linewidth}
\begin{tabular}{lcccc}
\toprule
Model & NFE=16 & NFE=32 & NFE=64 & NFE=128 \\
\midrule
ConvergeFlow & $4.3596\pm0.0013$ & $4.3398\pm0.0010$ & $4.3279\pm0.0011$ & $4.3195\pm0.0010$ \\
LangFlow (trainable $E$) & $4.3701\pm0.0013$ & $4.3523\pm0.0028$ & $4.3402\pm0.0037$ & $4.3328\pm0.0019$ \\
LangFlow (frozen $E$) & $4.3698\pm0.0008$ & $4.3514\pm0.0014$ & $4.3404\pm0.0011$ & $4.3328\pm0.0009$ \\
TIBF (trainable $E$) & $4.3554\pm0.0018$ & $4.3356\pm0.0017$ & $4.3231\pm0.0025$ & $4.3151\pm0.0024$ \\
TIBF & $4.3524\pm0.0004$ & $4.3348\pm0.0011$ & $4.3231\pm0.0013$ & $4.3143\pm0.0002$ \\
\bottomrule
\end{tabular}
\end{adjustbox}
\endgroup
\end{minipage}\par\smallskip

\par\smallskip\noindent\begin{minipage}{\linewidth}
\textbf{Repeated 4-grams (\%)}\par\smallskip
\begingroup
\setlength{\tabcolsep}{2pt}
\scriptsize
\begin{adjustbox}{max width=\linewidth}
\begin{tabular}{lcccc}
\toprule
Model & NFE=16 & NFE=32 & NFE=64 & NFE=128 \\
\midrule
ConvergeFlow & $\boldsymbol{0.0484\pm0.0109}$ & $0.0911\pm0.0097$ & $0.1388\pm0.0152$ & $0.1802\pm0.0138$ \\
LangFlow (trainable $E$) & $0.0505\pm0.0058$ & $0.0958\pm0.0189$ & $0.1461\pm0.0219$ & $0.2042\pm0.0235$ \\
LangFlow (frozen $E$) & $0.0570\pm0.0068$ & $0.1039\pm0.0211$ & $0.1581\pm0.0131$ & $0.1990\pm0.0043$ \\
TIBF (trainable $E$) & $0.0547\pm0.0118$ & $\boldsymbol{0.0891\pm0.0109}$ & $\boldsymbol{0.1255\pm0.0046}$ & $\boldsymbol{0.1622\pm0.0154}$ \\
TIBF & $0.0529\pm0.0037$ & $0.0896\pm0.0102$ & $0.1331\pm0.0212$ & $0.1758\pm0.0039$ \\
\bottomrule
\end{tabular}
\end{adjustbox}
\endgroup
\end{minipage}\par\smallskip

\par\medskip\noindent\begin{minipage}{\linewidth}
\refstepcounter{figure}\label{fig:owt_budget_entropy}
\textbf{Figure \thefigure: Gen.PPL and token entropy across sampling budgets.} Left: OWT; right: LM1B. Curves connect increasing NFE. Points are three-seed means; error bars show sample standard deviations.

\begin{center}
\includegraphics[width=\linewidth]{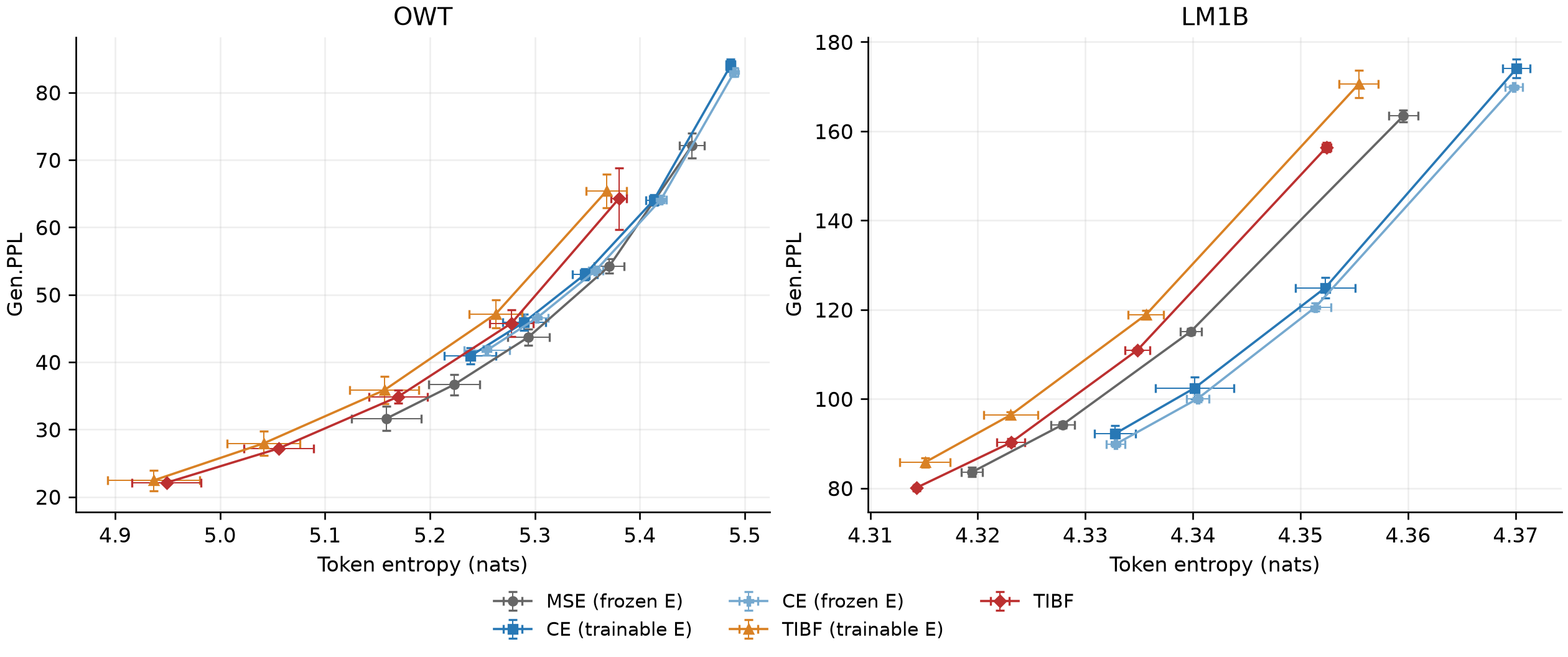}
\end{center}
\end{minipage}\par\medskip

\FloatBarrier
\subsection{Temperature Sensitivity}\label{sec:app:F.temperature}

Temperature $T$ scales the complete candidate scores after combining network and Gaussian observation scores. At each position, using the total score $s_v$ defined in the main text, the adjusted probabilities are
\begin{equation}
p_v(T)=\frac{\exp(s_v/T)}{\sum_{j=1}^V\exp(s_j/T)},\qquad T>0.
\tag{F.4}
\label{eq:app:F.added26}
\end{equation}
Generation uses $p(T)$ to weight embeddings and form a continuous prediction. All five models use $T\in\{0.8,0.9,1.0,1.1,1.2\}$, where $T=1$ is the primary setting. Temperatures share initial noise and sampling grids; other settings follow primary evaluation. Results appear in Tables~\ref{tab:temperature_owt} and \ref{tab:temperature_lm1b}.

\par\medskip\noindent\begin{minipage}{\linewidth}
\refstepcounter{figure}\label{fig:temperature_sensitivity}
\textbf{Figure \thefigure: Temperature sensitivity at fixed NFE.} The top row plots Gen.PPL against token entropy, and the bottom row against repeated 4-gram fraction. Filled markers denote $T=1$; open markers denote other temperatures. Lines connect three-seed means at the five prespecified temperatures; error bars show sample standard deviations.

\begin{center}
\includegraphics[width=\linewidth]{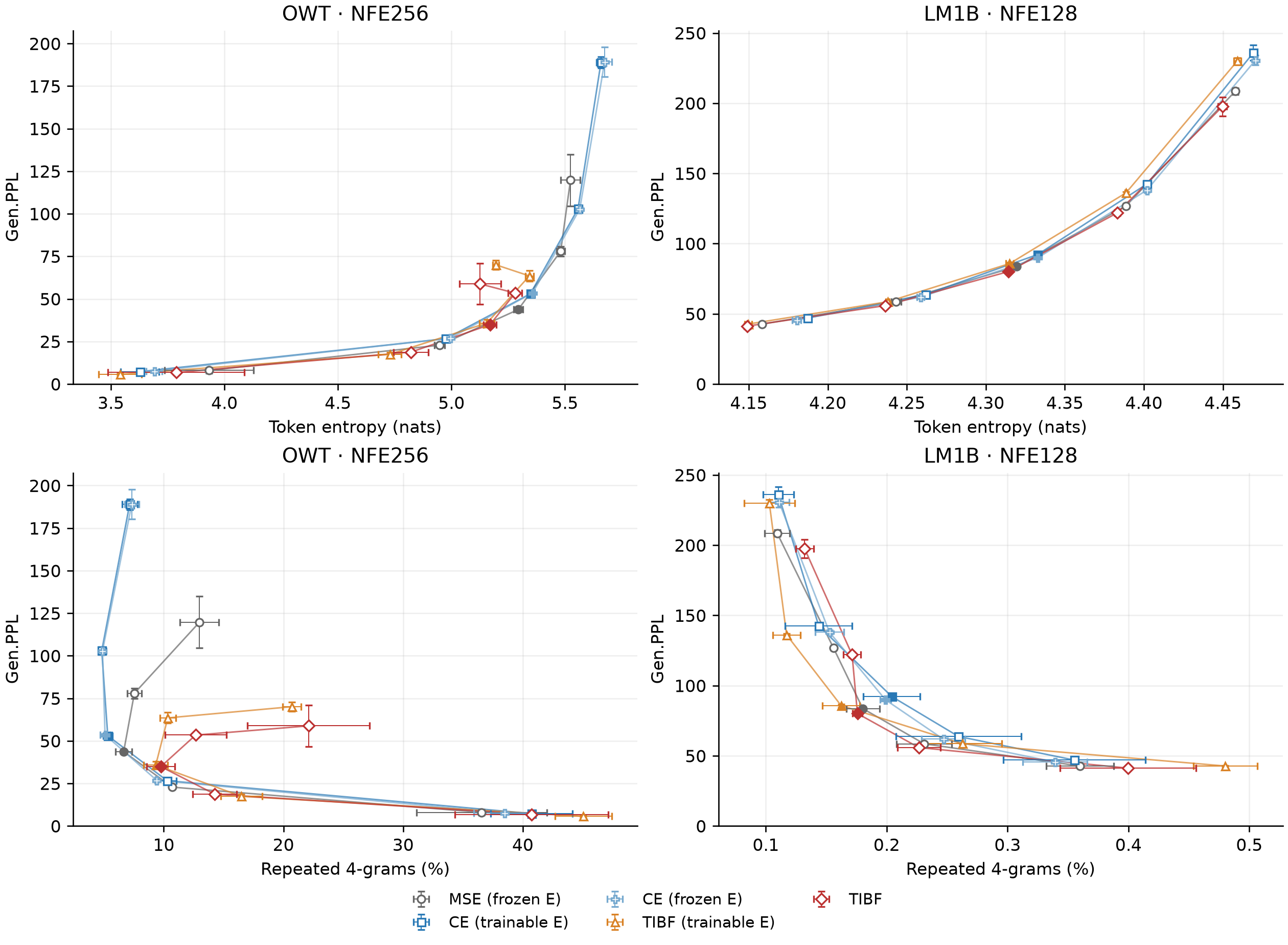}
\end{center}
\end{minipage}\par\medskip

\refstepcounter{table}\label{tab:temperature_owt}
\textbf{Table \thetable: OWT results at prespecified temperatures.} NFE256 and length 1,024 are fixed. Entries are means $\pm$ sample SD across three training seeds; repeated 4-grams are percentages.

\par\smallskip\noindent\begin{minipage}{\linewidth}
\textbf{Gen.PPL}

\begingroup
\setlength{\tabcolsep}{2pt}
\scriptsize
\begin{adjustbox}{max width=\linewidth}
\begin{tabular}{lccccc}
\toprule
Model & $T=0.8$ & $T=0.9$ & $T=1.0$ & $T=1.1$ & $T=1.2$ \\ 
\midrule
ConvergeFlow & $7.99\pm1.72$ & $22.82\pm0.70$ & $43.71\pm1.23$ & $77.95\pm2.97$ & $119.70\pm15.16$ \\
LangFlow (trainable $E$) & $7.26\pm0.66$ & $26.61\pm0.64$ & $53.03\pm0.84$ & $102.97\pm2.15$ & $188.91\pm3.26$ \\
LangFlow (frozen $E$) & $7.60\pm0.50$ & $26.92\pm0.51$ & $53.59\pm0.46$ & $102.75\pm1.31$ & $189.11\pm8.70$ \\
TIBF (trainable $E$) & $\boldsymbol{5.68\pm0.53}$ & $\boldsymbol{17.41\pm1.23}$ & $35.84\pm2.07$ & $63.49\pm3.17$ & $70.07\pm2.73$ \\
TIBF & $6.78\pm1.30$ & $18.74\pm1.35$ & $\boldsymbol{34.87\pm0.98}$ & $\boldsymbol{53.40\pm1.35}$ & $\boldsymbol{58.81\pm12.04}$ \\
\bottomrule
\end{tabular}
\end{adjustbox}
\endgroup
\end{minipage}\par\smallskip

\par\smallskip\noindent\begin{minipage}{\linewidth}
\textbf{Token entropy}

\begingroup
\setlength{\tabcolsep}{2pt}
\scriptsize
\begin{adjustbox}{max width=\linewidth}
\begin{tabular}{lccccc}
\toprule
Model & $T=0.8$ & $T=0.9$ & $T=1.0$ & $T=1.1$ & $T=1.2$ \\ 
\midrule
ConvergeFlow & $3.9324\pm0.1963$ & $4.9478\pm0.0218$ & $5.2939\pm0.0198$ & $5.4810\pm0.0182$ & $5.5242\pm0.0423$ \\
LangFlow (trainable $E$) & $3.6275\pm0.0843$ & $4.9753\pm0.0168$ & $5.3477\pm0.0119$ & $5.5575\pm0.0058$ & $5.6585\pm0.0200$ \\
LangFlow (frozen $E$) & $3.6906\pm0.0353$ & $4.9939\pm0.0048$ & $5.3580\pm0.0070$ & $5.5651\pm0.0079$ & $5.6744\pm0.0322$ \\
TIBF (trainable $E$) & $3.5412\pm0.0948$ & $4.7285\pm0.0506$ & $5.1566\pm0.0327$ & $5.3432\pm0.0181$ & $5.1947\pm0.0130$ \\
TIBF & $3.7871\pm0.3010$ & $4.8216\pm0.0768$ & $5.1699\pm0.0279$ & $5.2803\pm0.0300$ & $5.1256\pm0.0913$ \\
\bottomrule
\end{tabular}
\end{adjustbox}
\endgroup
\end{minipage}\par\smallskip

\par\smallskip\noindent\begin{minipage}{\linewidth}
\textbf{Distinct-2}

\begingroup
\setlength{\tabcolsep}{2pt}
\scriptsize
\begin{adjustbox}{max width=\linewidth}
\begin{tabular}{lccccc}
\toprule
Model & $T=0.8$ & $T=0.9$ & $T=1.0$ & $T=1.1$ & $T=1.2$ \\ 
\midrule
ConvergeFlow & $\boldsymbol{0.4157\pm0.0503}$ & $0.6849\pm0.0040$ & $0.7676\pm0.0085$ & $0.7847\pm0.0079$ & $0.7359\pm0.0169$ \\
LangFlow (trainable $E$) & $0.3832\pm0.0150$ & $0.6995\pm0.0080$ & $0.7903\pm0.0047$ & $\boldsymbol{0.8240\pm0.0020}$ & $\boldsymbol{0.8091\pm0.0085}$ \\
LangFlow (frozen $E$) & $0.3921\pm0.0064$ & $\boldsymbol{0.7070\pm0.0015}$ & $\boldsymbol{0.7941\pm0.0038}$ & $0.8235\pm0.0042$ & $0.8087\pm0.0088$ \\
TIBF (trainable $E$) & $0.3287\pm0.0194$ & $0.6093\pm0.0166$ & $0.7205\pm0.0115$ & $0.7384\pm0.0078$ & $0.6341\pm0.0064$ \\
TIBF & $0.3761\pm0.0583$ & $0.6382\pm0.0197$ & $0.7224\pm0.0098$ & $0.7158\pm0.0217$ & $0.6231\pm0.0443$ \\
\bottomrule
\end{tabular}
\end{adjustbox}
\endgroup
\end{minipage}\par\smallskip

\par\smallskip\noindent\begin{minipage}{\linewidth}
\textbf{Repeated 4-grams (\%)}

\begingroup
\setlength{\tabcolsep}{2pt}
\scriptsize
\begin{adjustbox}{max width=\linewidth}
\begin{tabular}{lccccc}
\toprule
Model & $T=0.8$ & $T=0.9$ & $T=1.0$ & $T=1.1$ & $T=1.2$ \\ 
\midrule
ConvergeFlow & $\boldsymbol{36.55\pm5.45}$ & $10.70\pm0.10$ & $6.65\pm0.67$ & $7.54\pm0.62$ & $12.98\pm1.61$ \\
LangFlow (trainable $E$) & $40.73\pm3.40$ & $10.28\pm0.78$ & $5.32\pm0.38$ & $\boldsymbol{4.82\pm0.19}$ & $\boldsymbol{7.15\pm0.64}$ \\
LangFlow (frozen $E$) & $38.47\pm2.58$ & $\boldsymbol{9.39\pm0.14}$ & $\boldsymbol{5.09\pm0.40}$ & $4.88\pm0.31$ & $7.33\pm0.61$ \\
TIBF (trainable $E$) & $45.06\pm2.38$ & $16.48\pm1.72$ & $9.33\pm1.01$ & $10.33\pm0.65$ & $20.70\pm0.77$ \\
TIBF & $40.71\pm6.41$ & $14.25\pm1.85$ & $9.75\pm1.18$ & $12.66\pm2.57$ & $22.09\pm5.10$ \\
\bottomrule
\end{tabular}
\end{adjustbox}
\endgroup
\end{minipage}\par\smallskip

\refstepcounter{table}\label{tab:temperature_lm1b}
\textbf{Table \thetable: LM1B results at prespecified temperatures.} NFE128 and length 128 are fixed. Entries are means $\pm$ sample SD across three training seeds; repeated 4-grams are percentages.

\par\smallskip\noindent\begin{minipage}{\linewidth}
\textbf{Gen.PPL}

\begingroup
\setlength{\tabcolsep}{2pt}
\scriptsize
\begin{adjustbox}{max width=\linewidth}
\begin{tabular}{lccccc}
\toprule
Model & $T=0.8$ & $T=0.9$ & $T=1.0$ & $T=1.1$ & $T=1.2$ \\ 
\midrule
ConvergeFlow & $42.69\pm0.11$ & $58.50\pm0.14$ & $83.63\pm1.01$ & $126.71\pm1.26$ & $208.52\pm2.31$ \\
LangFlow (trainable $E$) & $46.95\pm0.36$ & $63.77\pm0.20$ & $92.21\pm1.84$ & $142.52\pm1.07$ & $235.99\pm5.46$ \\
LangFlow (frozen $E$) & $45.55\pm0.38$ & $61.82\pm0.42$ & $89.87\pm0.60$ & $138.02\pm1.52$ & $230.60\pm3.46$ \\
TIBF (trainable $E$) & $42.71\pm0.46$ & $58.57\pm0.25$ & $85.79\pm1.03$ & $136.09\pm0.96$ & $230.05\pm2.37$ \\
TIBF & $\boldsymbol{41.21\pm0.56}$ & $\boldsymbol{55.86\pm0.25}$ & $\boldsymbol{80.08\pm0.63}$ & $\boldsymbol{121.96\pm2.09}$ & $\boldsymbol{197.57\pm6.61}$ \\
\bottomrule
\end{tabular}
\end{adjustbox}
\endgroup
\end{minipage}\par\smallskip

\par\smallskip\noindent\begin{minipage}{\linewidth}
\textbf{Token entropy}

\begingroup
\setlength{\tabcolsep}{2pt}
\scriptsize
\begin{adjustbox}{max width=\linewidth}
\begin{tabular}{lccccc}
\toprule
Model & $T=0.8$ & $T=0.9$ & $T=1.0$ & $T=1.1$ & $T=1.2$ \\ 
\midrule
ConvergeFlow & $4.1584\pm0.0008$ & $4.2433\pm0.0030$ & $4.3195\pm0.0010$ & $4.3885\pm0.0014$ & $4.4578\pm0.0017$ \\
LangFlow (trainable $E$) & $4.1871\pm0.0006$ & $4.2618\pm0.0005$ & $4.3328\pm0.0019$ & $4.4020\pm0.0022$ & $4.4693\pm0.0021$ \\
LangFlow (frozen $E$) & $4.1804\pm0.0027$ & $4.2586\pm0.0022$ & $4.3328\pm0.0009$ & $4.4018\pm0.0011$ & $4.4705\pm0.0011$ \\
TIBF (trainable $E$) & $4.1497\pm0.0023$ & $4.2378\pm0.0011$ & $4.3151\pm0.0024$ & $4.3886\pm0.0007$ & $4.4591\pm0.0023$ \\
TIBF & $4.1490\pm0.0020$ & $4.2365\pm0.0014$ & $4.3143\pm0.0002$ & $4.3833\pm0.0008$ & $4.4496\pm0.0029$ \\
\bottomrule
\end{tabular}
\end{adjustbox}
\endgroup
\end{minipage}\par\smallskip

\par\smallskip\noindent\begin{minipage}{\linewidth}
\textbf{Distinct-2}

\begingroup
\setlength{\tabcolsep}{2pt}
\scriptsize
\begin{adjustbox}{max width=\linewidth}
\begin{tabular}{lccccc}
\toprule
Model & $T=0.8$ & $T=0.9$ & $T=1.0$ & $T=1.1$ & $T=1.2$ \\ 
\midrule
ConvergeFlow & $0.9290\pm0.0004$ & $0.9430\pm0.0008$ & $0.9531\pm0.0002$ & $0.9612\pm0.0005$ & $0.9689\pm0.0006$ \\
LangFlow (trainable $E$) & $\boldsymbol{0.9342\pm0.0014}$ & $\boldsymbol{0.9460\pm0.0007}$ & $\boldsymbol{0.9552\pm0.0010}$ & $0.9632\pm0.0006$ & $\boldsymbol{0.9705\pm0.0003}$ \\
LangFlow (frozen $E$) & $0.9326\pm0.0007$ & $0.9447\pm0.0005$ & $0.9548\pm0.0002$ & $\boldsymbol{0.9633\pm0.0002}$ & $\boldsymbol{0.9705\pm0.0004}$ \\
TIBF (trainable $E$) & $0.9261\pm0.0006$ & $0.9419\pm0.0009$ & $0.9531\pm0.0008$ & $0.9620\pm0.0004$ & $0.9696\pm0.0006$ \\
TIBF & $0.9269\pm0.0010$ & $0.9421\pm0.0006$ & $0.9526\pm0.0007$ & $0.9605\pm0.0003$ & $0.9678\pm0.0005$ \\
\bottomrule
\end{tabular}
\end{adjustbox}
\endgroup
\end{minipage}\par\smallskip

\par\smallskip\noindent\begin{minipage}{\linewidth}
\textbf{Repeated 4-grams (\%)}

\begingroup
\setlength{\tabcolsep}{2pt}
\scriptsize
\begin{adjustbox}{max width=\linewidth}
\begin{tabular}{lccccc}
\toprule
Model & $T=0.8$ & $T=0.9$ & $T=1.0$ & $T=1.1$ & $T=1.2$ \\ 
\midrule
ConvergeFlow & $0.3599\pm0.0278$ & $0.2310\pm0.0228$ & $0.1802\pm0.0138$ & $0.1560\pm0.0018$ & $0.1094\pm0.0102$ \\
LangFlow (trainable $E$) & $0.3552\pm0.0588$ & $0.2594\pm0.0519$ & $0.2042\pm0.0235$ & $0.1438\pm0.0277$ & $0.1104\pm0.0127$ \\
LangFlow (frozen $E$) & $\boldsymbol{0.3391\pm0.0267}$ & $0.2469\pm0.0180$ & $0.1990\pm0.0043$ & $0.1526\pm0.0119$ & $0.1109\pm0.0082$ \\
TIBF (trainable $E$) & $0.4802\pm0.0263$ & $0.2625\pm0.0328$ & $\boldsymbol{0.1622\pm0.0154}$ & $\boldsymbol{0.1172\pm0.0116}$ & $\boldsymbol{0.1029\pm0.0209}$ \\
TIBF & $0.3995\pm0.0562$ & $\boldsymbol{0.2266\pm0.0177}$ & $0.1758\pm0.0039$ & $0.1714\pm0.0071$ & $0.1320\pm0.0075$ \\
\bottomrule
\end{tabular}
\end{adjustbox}
\endgroup
\end{minipage}\par\smallskip

TIBF achieves the lowest mean Gen.PPL at all five temperatures on LM1B and at $T\ge1$ on OWT; TIBF (trainable $E$) scores lower at $T<1$ on OWT. Reducing temperature below the default value of 1 lowers Gen.PPL but also increases repetition: on OWT, TIBF goes from Gen.PPL 34.87 and a repetition rate of 9.75\% at $T=1$ to 6.78 and 40.71\% at $T=0.8$. Temperature sensitivity therefore reflects a quality--diversity trade-off rather than an improvement in every metric.

\FloatBarrier
\subsection{Paired Confidence Intervals for Gen.PPL Differences}\label{sec:app:F.paired}

We compare mean Gen.PPL differences across the three training seeds using the shared initial-noise sequence indices from primary evaluation. Each bootstrap replicate samples 1,024 indices with replacement and applies them jointly to all models and training seeds. For each checkpoint, NLL is aggregated over valid scored tokens and exponentiated to obtain Gen.PPL. Scores are then averaged arithmetically across training seeds, and model differences are computed.

Per-sequence NLLs are calibrated within each original scoring batch so that their token-weighted sum equals the batch total NLL. With RNG seed 20261024, we perform 10,000 replicates and take the 2.5th and 97.5th percentiles as the 95\% interval. This interval measures generated-sample variability conditional on the three fixed sets of trained checkpoints. Results appear in Tables~\ref{tab:paired_owt} and \ref{tab:paired_lm1b}.

\begin{table}[!htbp]
\centering
\caption{Three-seed mean Gen.PPL differences at OWT NFE256. Each lower-triangular cell reports the row model minus the column model, with its paired 95\% interval; negative values indicate lower mean Gen.PPL for the row model.}
\label{tab:paired_owt}
\begingroup
\footnotesize
\setlength{\tabcolsep}{3pt}
\renewcommand{\arraystretch}{1.15}
\begin{adjustbox}{max width=\linewidth}
\begin{tabular}{lccccc}
\toprule
Row / column model & ConvergeFlow & \shortstack[c]{LangFlow\\(trainable $E$)} & \shortstack[c]{LangFlow\\(frozen $E$)} & \shortstack[c]{TIBF\\(trainable $E$)} & TIBF \\
\midrule
ConvergeFlow & --- &  &  &  &  \\[4pt]
\shortstack[l]{LangFlow\\(trainable $E$)} & \shortstack[c]{$9.3164$\\$[8.6573,$\\$9.9693]$} & --- &  &  &  \\[4pt]
\shortstack[l]{LangFlow\\(frozen $E$)} & \shortstack[c]{$9.8770$\\$[9.2724,$\\$10.4900]$} & \shortstack[c]{$0.5606$\\$[-0.1060,$\\$1.2427]$} & --- &  &  \\[4pt]
\shortstack[l]{TIBF\\(trainable $E$)} & \shortstack[c]{$-7.8677$\\$[-8.6320,$\\$-7.0852]$} & \shortstack[c]{$-17.1841$\\$[-18.0324,$\\$-16.3217]$} & \shortstack[c]{$-17.7447$\\$[-18.6580,$\\$-16.8676]$} & --- &  \\[4pt]
TIBF & \shortstack[c]{$-8.8357$\\$[-9.5085,$\\$-8.1820]$} & \shortstack[c]{$-18.1521$\\$[-18.9784,$\\$-17.3386]$} & \shortstack[c]{$-18.7127$\\$[-19.5426,$\\$-17.8977]$} & \shortstack[c]{$-0.9680$\\$[-1.7751,$\\$-0.1663]$} & --- \\[4pt]
\bottomrule
\end{tabular}
\end{adjustbox}
\endgroup
\end{table}

\begin{table}[!htbp]
\centering
\caption{Three-seed mean Gen.PPL differences at LM1B NFE128. Each lower-triangular cell reports the row model minus the column model, with its paired 95\% interval; negative values indicate lower mean Gen.PPL for the row model.}
\label{tab:paired_lm1b}
\begingroup
\footnotesize
\setlength{\tabcolsep}{3pt}
\renewcommand{\arraystretch}{1.15}
\begin{adjustbox}{max width=\linewidth}
\begin{tabular}{lccccc}
\toprule
Row / column model & ConvergeFlow & \shortstack[c]{LangFlow\\(trainable $E$)} & \shortstack[c]{LangFlow\\(frozen $E$)} & \shortstack[c]{TIBF\\(trainable $E$)} & TIBF \\
\midrule
ConvergeFlow & --- &  &  &  &  \\[4pt]
\shortstack[l]{LangFlow\\(trainable $E$)} & \shortstack[c]{$8.5799$\\$[7.1565,$\\$10.0478]$} & --- &  &  &  \\[4pt]
\shortstack[l]{LangFlow\\(frozen $E$)} & \shortstack[c]{$6.2463$\\$[5.2946,$\\$7.1855]$} & \shortstack[c]{$-2.3335$\\$[-3.8624,$\\$-0.8324]$} & --- &  &  \\[4pt]
\shortstack[l]{TIBF\\(trainable $E$)} & \shortstack[c]{$2.1652$\\$[0.7310,$\\$3.6237]$} & \shortstack[c]{$-6.4147$\\$[-7.8181,$\\$-5.0494]$} & \shortstack[c]{$-4.0811$\\$[-5.6575,$\\$-2.4781]$} & --- &  \\[4pt]
TIBF & \shortstack[c]{$-3.5469$\\$[-4.3810,$\\$-2.7378]$} & \shortstack[c]{$-12.1268$\\$[-13.5348,$\\$-10.7538]$} & \shortstack[c]{$-9.7932$\\$[-10.8114,$\\$-8.7797]$} & \shortstack[c]{$-5.7121$\\$[-7.1341,$\\$-4.3163]$} & --- \\[4pt]
\bottomrule
\end{tabular}
\end{adjustbox}
\endgroup
\end{table}

\FloatBarrier

On both datasets, the intervals for the three-seed mean Gen.PPL differences between TIBF and continuous MSE, either CE control, or TIBF (trainable $E$) lie below zero. The mean differences relative to TIBF (trainable $E$) are $-0.9680$ on OWT and $-5.7121$ on LM1B. These are pairwise 95\% intervals without adjustment for multiple comparisons.

\FloatBarrier
\subsection{Zero-Shot Cross-Corpus Transfer}\label{sec:app:F.transfer}
The five OWT models use their final EMA checkpoints from training seeds 17, 42, and 1234, without updating parameters on target corpora. PTB, WikiText-103-raw-v1, PubMed, and Arxiv use validation splits; PubMed and Arxiv use the article field of scientific\_papers. Models and training seeds share tokenization, blocking, and evaluation samples.

Text is restored using corpus-specific detokenization rules and encoded with the GPT-2 vocabulary. Within each group of 1,000 raw records, EOS is appended to each record before concatenation; content is split into 1,022-token blocks and wrapped with BOS and EOS to obtain length 1,024, discarding incomplete tails. For corpora with more than 1,024 blocks, NumPy PCG64 with seed20261029 selects 1,024 blocks without replacement, sorted by original block index; smaller corpora use all blocks. Evaluation uses the detached protocol in Appendix~\ref{sec:app:E.4}, with evaluation seed 1, 128 Heun-2 steps, and batch size 2. Results appear in Table~\ref{tab:owt_zeroshot}.

\begin{table}[htbp]
\centering
\caption{Zero-shot transfer dataset PPL of OWT continuation models. Parentheses give sequence counts.}
\label{tab:owt_zeroshot}
\begin{adjustbox}{max width=\linewidth}
\begin{tabular}{lcccc}
\toprule
Model & \shortstack{PTB\\(77)} & \shortstack{WikiText\\(232)} & \shortstack{PubMed\\(1024)} & \shortstack{Arxiv\\(1024)} \\
\midrule
ConvergeFlow & $155.50\pm4.95$ & $\boldsymbol{45.54\pm0.89}$ & $\boldsymbol{51.47\pm0.73}$ & $\boldsymbol{42.24\pm0.74}$ \\
LangFlow (trainable $E$) & $173.34\pm5.26$ & $48.84\pm1.30$ & $56.78\pm0.57$ & $47.38\pm0.80$ \\
LangFlow (frozen $E$) & $165.30\pm3.17$ & $48.18\pm1.23$ & $54.48\pm0.39$ & $45.30\pm0.26$ \\
TIBF (trainable $E$) & $178.59\pm0.84$ & $53.08\pm1.97$ & $60.15\pm0.38$ & $48.87\pm0.39$ \\
TIBF & $\boldsymbol{148.80\pm2.10}$ & $46.36\pm1.36$ & $\boldsymbol{51.47\pm0.93}$ & $43.10\pm0.43$ \\
\bottomrule
\end{tabular}
\end{adjustbox}
\end{table}

TIBF has the lowest mean PPL on PTB and matches continuous MSE closely on PubMed, while MSE scores lower on WikiText and Arxiv. TIBF outperforms both CE controls and TIBF (trainable $E$) on all four corpora, showing that retaining pretrained embeddings also improves cross-corpus modeling relative to these continuation settings.

\FloatBarrier
\clearpage
\section{Trajectory Diagnostics with Fixed Pretrained Embeddings}\label{sec:app:G}

\subsection{Diagnostic Settings and Distance Measures}\label{sec:app:G.1}

Diagnostics use the final 100,000-step EMA of OWT TIBF trained with seed 17. Sixteen generated sequences of length 1,024 provide 16,384 positions. Generation uses GPU seed 42 and batch size 8. Throughout, $\mathrm{logSNR}=-\gamma$. The main text reports candidate-distance separation and decoding agreement; here we supplement them with tail-distance contraction toward fixed targets and recovery after local perturbations.

Let $z_{i,s,\ell}$ be the state at node $i$, sequence $s$, and position $\ell$, with path coefficients $\alpha_i,\sigma_i$, embedding vector $e_j$, and dimension $d$. Define

\begin{equation}
U_{i,s,\ell}=z_{i,s,\ell}/\alpha_i,\quad h_i=\sigma_i/\alpha_i,\quad D_{i,s,\ell,j}=\frac{\|U_{i,s,\ell}-e_j\|}{\sqrt d\,h_i}.\tag{G.1}\label{eq:app:G.1}
\end{equation}

Sorting over the entire vocabulary yields nearest and second-nearest distances. The central curves show the midpoints of the 1\%--99\% quantile intervals across 16,384 positions; shading shows those intervals. The 95\% decoding-agreement intervals use 10,000 complete-sequence bootstrap replicates over the 16 sequences, with seed 20261022. The three decoding rules are probability argmax, state nearest neighbor, and continuous-prediction nearest neighbor.

The trajectory grid is uniform in $t=\alpha/(\alpha+\sigma)$, with logSNR endpoints $-8$ and 5. The distance curve uses NFE256; decoding agreement compares NFE32, 64, 128, 256, and 512. Inference uses the BF16 backbone, FP32 states, and FP64 Euler coefficients. Distance searches disable TF32; selected-candidate FP64 residuals and independent full-vocabulary FP64 subset checks verify distances and token choices.

\subsection{Endpoint Stability under Continued Denoising}\label{sec:app:G.2}

At each position, fix the probability-argmax token $j_*$ at $\log\mathrm{SNR}=5$ as the measurement target. Extend the same state and self-conditioning cache to logSNR20, using uniform logSNR steps of 1, 0.5, and 0.25. Each step recomputes token probabilities and updates the cache. TF32 is disabled to reduce low-noise rounding error. Figure~\ref{fig:endpoint_stability} reports the mean distance $d_{\mathrm{tar}}=\|U-e_{j_*}\|/\sqrt d$ and target-retention rates under the three decoding rules.

\par\medskip\noindent\begin{minipage}{\linewidth}
\refstepcounter{figure}\label{fig:endpoint_stability}
\textbf{Figure \thefigure: Endpoint distance and identity stability with fixed embeddings.} Left: mean $d_{\mathrm{tar}}$ for three grids on a logarithmic vertical axis; the dashed reference $d_{\mathrm{tar}}\propto h$ passes through the initial point. Right: percentage of positions decoded as $j_*$ on the finest grid. Probability argmax and nearest neighbor of the denoised prediction coincide at the recorded nodes.

\begin{center}
\includegraphics[width=\linewidth]{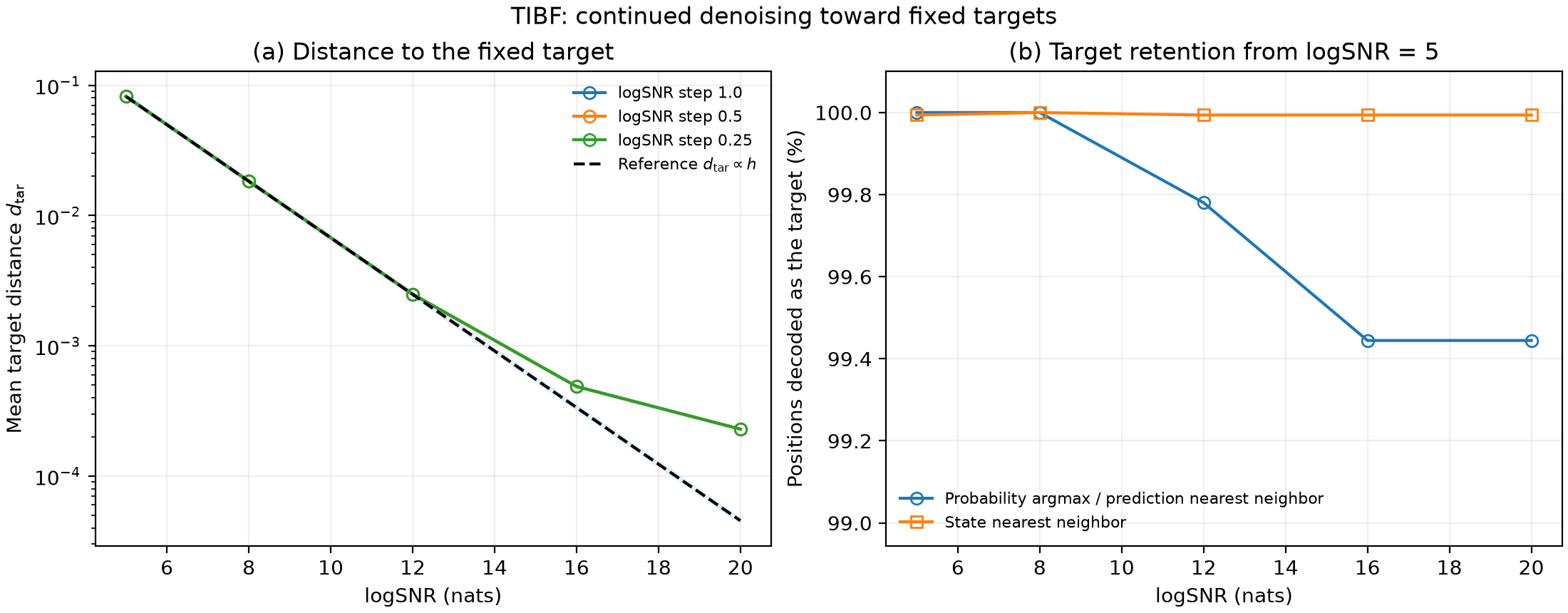}
\end{center}
\end{minipage}\par\medskip

On the finest grid, mean distance to the fixed target decreases from 0.08210 to $2.289\times10^{-4}$, a decrease by a factor of approximately 359. The minimum recorded identity-retention rate across the three decoders is 99.44\%. Across the three grids, a total of 421 probability-argmax token switches occur between consecutive prediction nodes.

\subsection{Recovery after Local Perturbations}\label{sec:app:G.3}

Full-vocabulary FP64 search gives minimum embedding separation $\Delta=1.86252$ and the radius specified by the theorem, $R=\Delta/4=0.465629$. Around the output embeddings above, apply two random-direction perturbations at each nominal radius $R/4,R/2,R$, giving 98,304 position instances. Initial radii are multiplied by 0.9999 to avoid rounding outside the boundary. Initialize self-conditioning with the target embedding, then advance the state using the actual network readouts and Euler updates. Local trajectories start at logSNR8 and use 16 updates of size 0.5 to reach logSNR16, for 17 network evaluations including the initial prediction. The two sets of random directions use GPU seeds 42 and 1042, incremented by the zero-based batch index. TF32 is disabled throughout the local trajectories.

\par\medskip\noindent\begin{minipage}{\linewidth}
\refstepcounter{figure}\label{fig:local_attraction}
\textbf{Figure \thefigure: Contraction after local embedding perturbations.} Left: group mean distance relative to initialization, with dashed $h/h_0$ reference, where $h_0$ is the initial noise-to-signal ratio of the local trajectory. Right: maximum Euclidean distance divided by $R$, with boundary at one.

\begin{center}
\includegraphics[width=\linewidth]{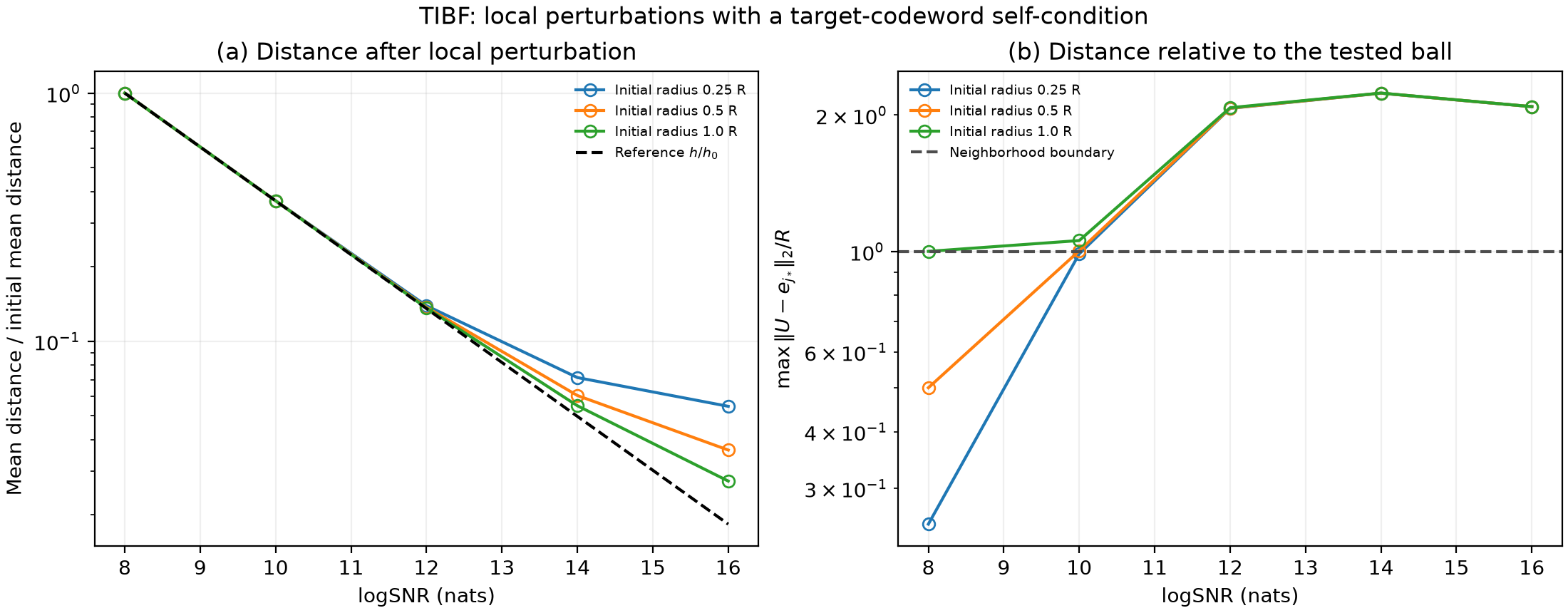}
\end{center}
\end{minipage}\par\medskip

\par\smallskip\noindent\begin{minipage}{\linewidth}
\refstepcounter{table}\label{tab:local_attraction}
\textbf{Table \thetable: Mean contraction and maximum deviation after local perturbations.}
\begingroup
\small
\setlength{\tabcolsep}{2pt}
\noindent\begin{tabular}{@{}p{0.19\linewidth}p{0.42\linewidth}p{0.33\linewidth}@{}}
\toprule
Initial radius & Final/initial mean distance (\%) & Maximum recorded distance/$R$ \\
\midrule
$R/4$ & 5.4653 & 2.2339 \\
$R/2$ & 3.6425 & 2.2343 \\
$R$ & 2.7314 & 2.2354 \\
\bottomrule
\end{tabular}
\endgroup
\end{minipage}\par\smallskip
Mean distances contract in all three groups, supporting average recovery near token endpoints. Maximum distances exceed $R$, indicating that this initial setting does not keep all positions within the neighborhood analyzed in Appendix~\ref{sec:app:C.5}.

\clearpage